\documentclass{article} 
\usepackage{iclr2027_conference,times}

\usepackage{amsmath,amsfonts,bm}

\def\eqref#1{equation~\ref{#1}}

\def\1{\bm{1}}

\DeclareMathAlphabet{\mathsfit}{\encodingdefault}{\sfdefault}{m}{sl}
\SetMathAlphabet{\mathsfit}{bold}{\encodingdefault}{\sfdefault}{bx}{n}

\usepackage{booktabs, multirow} 
\usepackage{amsfonts}       
\usepackage{nicefrac}       
\usepackage{microtype}      
\usepackage{xcolor}         
\usepackage{hyperref}
\usepackage{url}
\usepackage{graphicx}
\usepackage{subcaption}
\usepackage{algorithm}%
\usepackage{algorithmic}%
\usepackage{setspace}
\usepackage{amsmath}
\usepackage{natbib}

\title{Learning Stock Trading Policies via Barycenter-Based Adversarial Inverse Reinforcement Learning}

\author{Arishi Orra, Himanshu Choudhary \& Manoj Thakur \\
School of Mathematical and Statistical Sciences\\
Indian Institue of Technology Mandi\\
Mandi, Himachal Pradesh, India, 175005 \\
\texttt{d21022@students.iitmandi.ac.in, ch.himanshu1199@gmail.com, manoj@iitmandi.ac.in}
}

\iclrfinalcopy 
\begin{document}

\maketitle

\begin{abstract}
  Designing effective trading strategies using reinforcement learning remains challenging due to delayed and noisy rewards, poor exploration, and the difficulty of enforcing explicit risk constraints. In this work, we propose BRaG, a barycenter-based adversarial inverse reinforcement learning framework for stock trading that learns trading behavior from multiple heterogeneous expert strategies. BRaG aggregates expert demonstrations using a performance-weighted Wasserstein barycenter, yielding a stable pseudo-expert representation that captures shared structure across diverse trading styles. This representation is used to pretrain a trading policy via adversarial imitation learning, which alleviates unstable exploration during reinforcement learning. The pretrained policy is subsequently refined using reinforcement learning with true market rewards. To ensure risk-aware decision-making, BRaG incorporates control barrier functions that constrain action execution and regularize policy learning to satisfy drawdown limits. We evaluate the proposed approach on four major global equity markets, including the US, UK, Indian, and Taiwanese indices. Across all the markets, the proposed approach achieves stronger performance than both classical trading rules and recent deep reinforcement learning methods, while exhibiting more stable risk characteristics. 
\end{abstract}

\section{Introduction} \label{Introduction}
Financial markets play a central role in modern economies by facilitating capital allocation, risk transfer, and price discovery. Within this broad landscape, stock trading has attracted sustained research interest owing to its economic relevance and the availability of large-scale historical data. At the same time, stock prices exhibit high volatility, non-stationarity, and sensitivity to exogenous events, which makes consistent decision-making challenging \citep{wu2020adaptive}. Traditional trading strategies based on fixed rules or handcrafted indicators often fail to adapt to evolving market conditions and may suffer from limited generalization. Supervised learning approaches have been used to predict future price movements or returns. However, their effectiveness is limited by label noise and the difficulty of translating predictions into profitable trading actions. As a result, there has been growing interest in learning frameworks that directly optimize sequential trading decisions through interaction with the market \citep{zhang2019deep, tsantekidis2020price, choudhary2026dynamic}.

Reinforcement learning (RL) has gained prominence as a natural framework for framing stock trading as a sequential decision-making problem. In a typical setting, an agent observes market states constructed from historical prices and related indicators, executes trading actions such as buy, sell, or hold, and receives feedback based on the resulting returns \citep{moody1998performance}. In contrast to supervised approaches, RL focuses on maximizing cumulative performance over time rather than short-term predictive accuracy. With the integration of deep neural networks, deep reinforcement learning (DRL) methods have been successfully applied to a wide range of trading tasks, including single-asset trading, portfolio management, and intraday trading. Existing studies indicate that DRL agents are capable of capturing temporal structure in financial time series and adjusting their behavior as market regimes evolve, which makes this paradigm well-suited for automated trading applications \citep{sun2022deepscalper, kabbani2022deep, choudhary2026cvar}.

Despite these advantages, deploying DRL in real-world stock trading remains non-trivial. One of the central challenges lies in designing reward functions that accurately capture desirable trading behavior over extended horizons. Market feedback is often sparse, delayed, and noisy, and reward signals based purely on short-term profits can lead to unstable learning dynamics \cite{choudhary2025risk}. In addition, manually specified rewards typically embed implicit assumptions about risk tolerance and market behavior, which may not transfer well across assets or changing conditions \citep{rodinos2023sharpe, orra2025enhancing}. While alternative objectives such as risk-adjusted returns or drawdown-based penalties can be incorporated, they require careful tuning and still provide only partial proxies for the complex preferences underlying successful trading strategies. Inverse reinforcement learning (IRL) offers a principled alternative by seeking to recover an underlying reward function that explains observed expert trajectories, under the assumption that the expert behaves optimally or near-optimally \citep{ng2000algorithms}. Recent adversarial formulations, such as Generative Adversarial Imitation Learning (GAIL), reformulate this problem as one of matching the distribution of expert and agent trajectories, enabling scalable policy learning without explicit reward specification \citep{ho2016generative}. Beyond reward recovery, this formulation also provides an effective pretraining mechanism, initializing the policy in a region of the policy space aligned with expert behavior and thereby mitigating the instability of exploration under sparse and noisy market rewards. In the context of stock trading, IRL provides a mechanism for capturing latent trading preferences reflected in expert decisions and for training agents whose behavior aligns more closely with expert-level strategies through interaction with the market \citep{sun2023transaction, roa2019towards, zhang2022reinforcement}.

Despite recent advances, important limitations remain in applying IRL to stock trading. Existing methods often rely on demonstrations from a single expert or a narrowly defined strategy, which is restrictive in markets characterized by heterogeneous and evolving trading behaviors. Such approaches are prone to overfitting and may fail to generalize across market regimes. Moreover, adversarial imitation learning methods are commonly applied without explicitly addressing the exploration challenges faced by reinforcement learning in noisy and high-dimensional financial environments. Policies trained from scratch frequently exhibit unstable learning dynamics due to sparse and delayed rewards \citep{liu2020adaptive, halperin2022combining}. In addition, most existing approaches either depend on manually specified reward functions or do not explicitly incorporate risk constraints during policy learning, limiting their ability to capture complex trading preferences and ensure stable performance. These limitations highlight a gap for a unified framework that can leverage diverse expert behaviors, avoid explicit reward specification, and incorporate risk awareness in a principled manner.

To address these challenges, we propose BRaG, a barycenter-based adversarial IRL framework for stock trading. BRaG aggregates multiple heterogeneous expert strategies using a performance-weighted Wasserstein barycenter, yielding a stable pseudo-expert distribution for imitation learning. This aggregated representation is used to pretrain a trading policy via adversarial imitation learning, which mitigates poor exploration by initializing the agent with expert-aligned behavior. The pretrained policy is subsequently refined using reinforcement learning with true market rewards. Explicit risk control is incorporated through control barrier functions that constrain action execution and regularize policy learning to ensure compliance with drawdown limits throughout training and deployment. We evaluate the proposed framework on four major global equity markets, including the US, UK, Indian, and Taiwanese stock indices. Empirical results demonstrate that BRaG consistently outperforms classical trading strategies and state-of-the-art DRL baselines in terms of both return and risk-adjusted performance. Although BRaG builds upon established techniques, its contribution lies in the principled integration of these techniques to address the practical challenges of multi-expert imitation learning for financial trading, including expert aggregation, stable policy initialization, adaptive policy refinement, and risk-aware optimization. The main contributions of this work are summarized as follows:

\begin{itemize}
    \item We propose BRaG, a barycenter-based adversarial IRL framework for stock trading that aggregates heterogeneous expert strategies to construct a stable pseudo-expert representation for imitation learning.
    
    \item We introduce a performance-weighted Wasserstein barycenter to combine multiple expert trajectories, thereby reducing sensitivity to individual expert bias and enhancing robustness across diverse market conditions.
    
    \item We demonstrate how adversarial imitation learning can be used as an effective pretraining mechanism to address poor exploration in reinforcement learning for financial markets.
    
    \item We integrate explicit risk control through control barrier functions, enforcing drawdown constraints at action execution time and regularizing policy learning during training.
    
    \item We conduct extensive empirical evaluation across four global equity markets, showing that the proposed approach consistently outperforms classical trading strategies and state-of-the-art DRL baselines in terms of both return and risk-adjusted performance.
\end{itemize}


\section{Preliminaries \label{Background}}

\paragraph{Reinforcement Learning.}
We model stock trading as a Markov Decision Process (MDP), where an agent observes market states, takes trading actions, and receives rewards based on portfolio performance. The objective is to learn a policy that maximizes expected cumulative return over time \citep{sutton1998reinforcement}. In this work, we adopt a policy gradient framework with actor–critic methods for stable optimization, specifically using Proximal Policy Optimization (PPO) \citep{schulman2017proximal}, with implementation details provided in Appendix~\ref{ppo}.

\paragraph{Inverse Reinforcement Learning.}
Inverse reinforcement learning (IRL) addresses the problem of inferring an underlying reward function from expert demonstrations, under the assumption that the expert behaves optimally or near-optimally. Rather than relying on manually specified objectives, IRL aims to recover latent preferences that explain observed behavior. This is particularly useful in financial domains, where reward design is challenging due to noisy, delayed, and non-stationary market feedback \citep{ng2000algorithms, fu2017learning}.

\paragraph{Adversarial Imitation Learning.}
Modern IRL approaches often adopt adversarial formulations that avoid explicit reward recovery. Generative Adversarial Imitation Learning (GAIL) casts policy learning as a distribution-matching problem between expert and agent trajectories. A discriminator is trained to distinguish expert samples from those generated by the policy, while the policy is optimized to produce behavior that is indistinguishable from expert demonstrations. The discriminator output serves as a surrogate reward signal, enabling scalable policy learning without manual reward specification. In our framework, this formulation is used to pretrain a trading policy that aligns with expert behavior before further refinement via reinforcement learning \citep{ho2016generative}.

\section{Problem Formulation \label{Problem}}
We study algorithmic stock trading as a sequential decision problem. At each trading time, an agent observes the market state, updates its portfolio, and receives feedback through profit and risk. This setting is naturally cast as an MDP, since the objective depends on a sequence of interlinked decisions and the reward is delayed and path‑dependent. Formulating the stock trading problem as an MDP enables the use of RL techniques to optimize trading strategies dynamically over time.

Let $\mathcal{I}=\{1,\dots,n\}$ be the list of tradable stocks and $t=\{0,1,\dots,T-1\}$ denote decision times. Then, for our multi-stock trading problem, the key components of the MDP are defined as follows:

\paragraph{State $s_t$:} The state collects market features and portfolio information. For an $n-$asset portfolio, the state comprises the available cash $b_t\in\mathbb{R}_+$, current share holdings $h_t\in\mathbb{Z}^n$, closing price of each stock $p_t \in \mathbb{R}_+^n$, and eight most commonly used technical indicators corresponding to each stock. This formulation assigns a compact dimension of $10n+1$ to the state $ s_t$ for $ n$ assets.

\paragraph{Action $a_t$:} The action \( a_t \) represents the trading decision across all stocks at time $t$. It is modeled as an $n-$dimensional vector $a_t = [a_{t,1}, \ldots, a_{t,n}]^\top$, where each component $a_{t,i} \in \{-m, \ldots, 0, \ldots, m\}$ denotes the number of shares of stock $i$ to be bought (positive), sold (negative), or held (zero). The parameter $m$ specifies the upper limit on the number of shares that can be traded in a single period. Consequently, the overall action space has a cardinality of $(2m + 1)^n$.

\paragraph{Reward $r_t$:} After executing the action $a_t$, the cash updates as
\[
b_{t+1}= b_t - p_t^\top a_t - c(a_t,p_t),
\]
where $c$ is the fixed commission fee. The reward $r_t$ is defined as the percentage change in the marked‑to‑market portfolio value, net of trading costs, 
\[
r_t = \frac{V_{t+1} - V_t}{V_t}, \qquad V_t = b_t + p_t^\top h_t.
\]
This reward formulation encourages the agent to pursue profit-maximizing trades while concurrently penalizing excessive transactions, thereby balancing return and cost efficiency.
 
\section{Methodology \label{Methodology}}
We consider the problem of learning an expert-level stock trading policy in settings where the underlying reward function governing expert behavior is unknown and multiple heterogeneous expert strategies are available. To address this, we propose BRaG (Barycenter-based Risk-aware Adversarial GAIL), a multi-expert adversarial inverse reinforcement learning framework for stock trading. The central idea behind BRaG is to leverage the diversity of multiple expert strategies while avoiding overfitting to any single trading style. This is achieved by aggregating heterogeneous expert demonstrations through a performance-weighted Wasserstein barycenter, which yields a stable and representative pseudo-expert distribution. Policy learning is then carried out using adversarial imitation learning, enabling implicit reward recovery without explicit reward specification. Finally, explicit portfolio risk constraints are enforced through control barrier functions during action execution.

\subsection{Multi-Expert Demonstrations and Occupancy Measures}

We assume access to $K$ expert trading strategies $\{\pi_k\}_{k=1}^K$, each reflecting a distinct trading philosophy. From each expert $\pi_k$, we observe a dataset of state-action pairs:
\begin{equation}
\mathcal{D}_k = \{(s_t^{(k)}, a_t^{(k)})\}_{t=1}^{T_k}.
\end{equation}

Each expert induces a discounted occupancy measure:
\begin{equation}
\rho_{\pi_k}(s,a) = \sum_{t=0}^{\infty} \gamma^t \Pr(s_t = s, a_t = a \mid \pi_k),
\end{equation}
which captures the long-term behavior of the expert. Rather than selecting a single expert or merging all demonstrations indiscriminately, BRaG seeks to construct a consensus representation that preserves common behavioral structure across experts.

\subsection{Performance-Weighted Barycenter of Expert Trajectories}

To aggregate heterogeneous expert behaviors into a single representative expert distribution, we compute a performance-weighted sliced Wasserstein barycenter over expert state-action trajectories. Unlike uniform mixing or single-expert selection, the Wasserstein barycenter preserves the geometry of expert behaviors and provides a principled consensus representation. This yields robustness to expert bias while capturing shared structure across heterogeneous strategies.

Each expert policy $\pi_k$ is assigned a non-negative weight $\alpha_k$ proportional to its risk-adjusted performance, measured using the Sharpe ratio:
\begin{equation}
\alpha_k =
\frac{\tilde S_k}{\sum_l \tilde S_l},
\end{equation}
where $\tilde S_k = \max\!\left( \mathrm{Sharpe}(\pi_k), 0 \right)$, the clipping at zero excludes experts with negative risk-adjusted performance from the aggregation. This weighting favors experts who exhibit superior return-to-risk characteristics. This Sharpe ratio-based weighting choice favors experts with high risk-adjusted performance rather than raw returns, so the barycenter is not skewed toward strategies that profited by taking on excessive risk.

Let $\mu_k$ denote the empirical distribution of concatenated state-action samples $(s,a)$ generated by expert $\pi_k$. The objective is to compute a barycenter distribution $\mu_{\mathrm{bar}}$ that minimizes the weighted sum of Wasserstein distances to individual expert distributions:
\begin{equation}
\mu_{\mathrm{bar}} = \arg\min_{\mu} \sum_{k=1}^K \alpha_k \, W_2^2(\mu, \mu_k),
\end{equation}
where $W_2$ denotes the 2-Wasserstein distance. 

Computing Wasserstein barycenters directly in high-dimensional spaces is intractable. We therefore adopt a sliced Wasserstein approximation. Specifically, random projection directions $v \in \mathbb{R}^{d_s + d_a}$ with $\|v\|_2 = 1$ are sampled, and expert samples are projected onto one-dimensional subspaces. For each projection, the barycenter is obtained by weighted averaging of the empirical quantile functions:
\begin{equation}
\hat{z}_{\mathrm{bar}}^{(v)}(q) = \sum_{k=1}^K \alpha_k \, \hat{z}_k^{(v)}(q),
\end{equation}
where $\hat{z}_k^{(v)}$ denotes the sorted projected samples of expert $k$. The barycenter quantile $\hat{z}_{\mathrm{bar}}^{(v)}(q)$ defines the projected barycenter distribution, from which pseudo-samples are obtained by sampling over quantile levels $q$. Since sliced Wasserstein barycenters do not admit a unique inverse mapping, we construct pseudo-expert samples by aggregating quantiles across multiple random projections.

The final barycenter samples are recovered by averaging over multiple random projections and back-projecting to the original space. The resulting dataset
\begin{equation}
\mathcal{D}_{\mathrm{bar}} = \{(s_t^{\mathrm{bar}}, a_t^{\mathrm{bar}})\}
\end{equation}
constitutes a set of pseudo-expert demonstrations. Intuitively, the barycenter aggregates expert trajectories in a structured manner rather than simple pooling. Expert datasets are first weighted by performance, and their state-action samples are projected onto multiple one-dimensional directions. For each projection, the barycenter is computed as a weighted average of the sorted projected samples (quantiles) across experts. Repeating this across projections yields pseudo-expert samples $\mathcal{D}_{\mathrm{bar}}$, which capture shared structure while reducing sensitivity to individual strategies. Since the sliced Wasserstein back projection does not admit a unique inverse map, a reconstructed sample may not correspond to a trade that is actually reachable given the agent's cash and holdings at that step. We address this by passing every reconstructed sample through a feasibility check before it enters $\mathcal{D}_{\mathrm{bar}}$. Each candidate action is compared against the available cash and current holdings, and is clipped to the nearest feasible point when a violation occurs.

\subsection{Adversarial Inverse Reinforcement Learning via GAIL}

BRaG employs Generative Adversarial Imitation Learning (GAIL) to recover an implicit reward function and a corresponding trading policy. GAIL frames imitation learning as a min-max game between a policy $\pi_\theta$ and a discriminator $D_\phi$.

The discriminator $D_\phi : \mathcal{S} \times \mathcal{A} \to (0,1)$ is trained to distinguish pseudo-expert samples from policy-generated samples by maximizing:
\begin{equation}
\mathbb{E}_{(s,a)\sim \rho_{\pi_\theta}}[\log D_\phi(s,a)]
+
\mathbb{E}_{(s,a)\sim \rho_{\mathrm{bar}}}[\log (1 - D_\phi(s,a))].
\end{equation}

From the discriminator, an implicit reward signal is obtained as follows: 
\begin{equation}
r_\phi(s,a) = -\log (D_\phi(s,a)).
\end{equation}
This reward assigns higher values to state-action pairs that resemble barycenter expert behavior. Unlike classical inverse reinforcement learning, this formulation avoids explicit reward parameterization and instead directly matches occupancy measures.

The policy $\pi_\theta$ is optimized to maximize the expected discounted sum of discriminator-derived rewards:
\begin{equation}
\max_\theta \;
\mathbb{E}_{\pi_\theta}\!\left[
\sum_{t=0}^{\infty} \gamma^t r_\phi(s_t,a_t)
\right].
\end{equation}
In BRaG, policy optimization is performed using Proximal Policy Optimization (PPO) \citep{schulman2017proximal}, which ensures stable updates through clipped likelihood ratios and value function regularization. This adversarial learning process iteratively aligns the policy distribution with the barycenter expert distribution.

\subsection{Risk-Constrained Action Selection using Control Barrier Functions}

Financial risk is explicitly controlled using a control barrier function (CBF) that enforces a maximum drawdown constraint. Let $V_t$ denote the portfolio value at time $t$, and let $V_t^{\max} = \max_{\tau \le t} V_\tau$ denote the running maximum portfolio value. The drawdown at time $t$ is defined as
\begin{equation}
\mathrm{DD}_t = 1 - \frac{V_t}{V_t^{\max}}.
\end{equation}
We define the barrier function:
\begin{equation}
B(V_t) = \delta_{\max} - \mathrm{DD}_t,
\end{equation}
where $\delta_{\max} \in (0,1)$ specifies the maximum allowable drawdown. Safety is ensured when $B(V_t) \ge 0$.

To account for uncertainty, we estimate a conservative lower-bound return vector using a rolling window:
\begin{equation}
r_t^{\mathrm{lb}} = \mu_t - \kappa \sigma_t,
\end{equation}
where $\mu_t, \sigma_t\in\mathbb{R}^n$ denote the rolling mean and standard deviation of asset returns up to time $t-1$, and $\kappa > 0$ controls risk sensitivity. Using this estimate, the next portfolio value $V_{t+1}$ is predicted conservatively as
\begin{equation}
    V_{t+1} = b_t - p_t^\top a_t - c(a_t,p_t) + \hat{p}^\top_{t+1} (h_t + a_t),
\end{equation}
where
\begin{equation}
    \hat{p}_{t+1} = p_t \odot (1+r_t^{\mathrm{lb}}).
\end{equation}

The CBF condition is enforced via:
\begin{equation}
B(V_{t+1}) - B(V_t) + \alpha_{\textbf{CBF}} B(V_t) \ge 0,
\end{equation}
where $\alpha_{\textbf{CBF}} > 0$ determines the rate at which the system is driven back towards the safe set.

If the condition is violated, the policy action $a_t$ is projected onto the safe set by solving
\begin{equation}
\begin{aligned}
a_t^{\mathrm{safe}} = \;& \arg\min_{a \in \mathcal{A}} \|a - a_t\|_2^2 \\
\text{s.t. } \;& B(V_{t+1}) - B(V_t) + \alpha_{\textbf{CBF}} B(V_t) \ge 0.
\end{aligned}
\end{equation}
In addition to action-level safety enforcement, we also incorporate a soft penalty term in the PPO objective to discourage repeated violations of the barrier condition during training. The combination of hard constraint enforcement at execution time and soft regularization during optimization improves training stability and reduces the need for frequent corrections.

\begin{algorithm}[!htp]
\caption{BRaG: Barycenter-based Risk-aware Adversarial GAIL}
\label{alg:brag}
\begin{algorithmic}[1]

\REQUIRE Expert datasets $\{\mathcal{D}_k\}_{k=1}^K$, parameters $\gamma,\delta_{\max},\kappa,\alpha$
\ENSURE Trained policy $\pi_\theta$

\STATE Compute expert weights $\alpha_k \propto \mathrm{Sharpe}(\pi_k)$
\STATE Compute sliced Wasserstein barycenter $\mathcal{D}_{\mathrm{bar}}$

\STATE Initialize policy $\pi_\theta$ and discriminator $D_\phi$

\WHILE{GAIL not converged}
    \STATE Collect rollouts $\tau_\pi$ using $\pi_\theta$
    \STATE Apply CBF-based action projection during rollouts
    \STATE Update $D_\phi$ using $\mathcal{D}_{\mathrm{bar}}$ and $\tau_\pi$
    \STATE Set $r_\phi(s,a)=-\log(D_\phi(s,a))$
    \STATE Update $\pi_\theta$ using PPO with reward $r_\phi$
\ENDWHILE

\WHILE{fine-tuning not converged}
    \STATE Roll out $\pi_\theta$ with true environment reward
    \STATE Enforce CBF constraints during action execution
    \STATE Update $\pi_\theta$ using PPO
\ENDWHILE

\RETURN $\pi_\theta$

\end{algorithmic}
\end{algorithm}

\subsection{Training Procedure}

Training in BRaG proceeds in a staged manner to ensure stability and effective knowledge transfer from expert demonstrations to the final trading policy.

In the first stage, the policy is initialized through adversarial imitation learning using the barycenter pseudo-expert demonstrations. A discriminator is updated to differentiate between policy-generated state-action pairs and barycenter expert samples. The output of the discriminator is used to compute an imitation-based reward signal, which replaces the environment reward during this phase. Policy updates are performed using PPO, allowing for stable improvement under the adversarial reward. Throughout this stage, every action proposed by the policy is filtered through the CBF before execution, ensuring that drawdown constraints are respected during data collection itself. This pretraining stage enables the policy to internalize expert-like trading behavior while operating within explicit risk limits.

In the second stage, the pretrained policy is further refined using reinforcement learning with the true environment reward. This phase enables the policy to adapt to market-specific dynamics and improve realized performance beyond imitation. The optimization objective is now aligned with portfolio returns, while the CBF continues to regulate action execution. By retaining the barrier mechanism during fine-tuning, risk constraints remain active even as the policy explores new behaviors. Model selection is based on validation performance, and the final policy is evaluated on an unseen trading period to assess generalization. The complete training procedure of BRaG is summarized in Algorithm~\ref{alg:brag}.

\section{Experimental Evaluation \label{Experiments}}
This section reports the experimental study carried out in this work. We first describe the market datasets used in the experiments and the corresponding experimental settings. The evaluation criteria and baseline methods are then introduced. The section concludes with the empirical results, along with a discussion of the observed performance across datasets.

\subsection{Data Description}

    The proposed methodology is evaluated using daily stock price data from four major global equity indices, representing diverse geographical regions and market structures. These include the Dow Jones Industrial Average (DJI) from the United States, the Financial Times Stock Exchange $100$ (FTSE $100$) from the United Kingdom, the BSE Sensex from India, and the Taiwan Capitalization Weighted Stock Index (TAIEX or TWII) from Taiwan. To maintain consistency across experiments, a fixed universe of thirty stocks is considered for each index. All constituent stocks of the DJI and Sensex are included. For the FTSE $100$ and TAIEX, the thirty stocks with the highest market capitalization are selected. Daily closing price data for all selected stocks is obtained from Yahoo Finance\footnote{\url{https://finance.yahoo.com/}} and spans the period from January $1$, $2010$, to September $30$, $2025$. These prices are used to compute technical indicators that form the state representation for the trading agent. The dataset for each index is divided into non-overlapping temporal segments to support training, validation, and out-of-sample evaluation. Data spanning January $1$, $2010$, to December $31$, $2020$, is used for GAIL-based pretraining, followed by fine-tuning on data from January $1$, $2021$, to December $31$, $2023$. The learned policies are then evaluated on an unseen test period ranging from January $1$, $2024$, to September $30$, $2025$.

\subsection{Experiment settings}

    All experiments are conducted under consistent environmental settings to ensure fair comparison across models. The trading setup operates at a daily frequency, where decisions are made on each trading day based on daily market features. At the start of each trading period, the agent is initialized with a capital of $1,000,000$. Transaction costs are modeled by applying a fee of $0.1\%$ on both buy and sell operations, and all trades are executed at the daily closing price. To capture heterogeneous expert behavior, five distinct expert trajectories are considered, including Cross-sectional momentum (CSMOM) \citep{jegadeesh2002cross}, Time-series momentum (TSMOM) \citep{moskowitz2012time}, Moving average (MA) crossover \citep{brock1992simple}, Bollinger Bands-based strategy \citep{kirkpatrick2010technical}, and a DRL-based PPO trading agent. These strategies are chosen to reflect complementary trading principles, including momentum, mean-reversion, trend-following, and learning-based approaches, thereby ensuring diversity in expert behavior rather than redundancy. A detailed description of the expert strategies is provided in Appendix \ref{expert}. Hyperparameters for the proposed approach and baseline DRL models are tuned using Bayesian optimization \citep{snoek2012practical} implemented through the Hyperopt framework. The search ranges are chosen based on prior empirical studies to strike a balance between performance and computational efficiency \citep{mnih2016asynchronous, schulman2017proximal}. The complete set of hyperparameters and their search ranges are reported in Appendix \ref{hyperparameter}. 

\subsection{Baselines and Performance Metrics}

    We compare BRaG against a broad set of baseline strategies spanning classical finance, heuristic trading rules, and deep reinforcement learning methods to ensure a comprehensive evaluation. Non-learning benchmarks include the market index, buy-and-hold, random trading, and the mean–variance optimization (MVO) model \citep{markowitz1990foundations}. Rule-based technical strategies consist of CSMOM, TSMOM, MA crossover, and the Bollinger Band strategy. We also evaluate learning-based baselines, including A2C \citep{mnih2016asynchronous}, PPO \citep{schulman2017proximal}, and DDPG \citep{lillicrap2015continuous}, along with five recent DRL-based trading approaches from the literature, namely, VS-DRL \citep{zhang2019deep}, SRRS \citep{rodinos2023sharpe}, RSHF \citep{orra2025enhancing}, Adaptive \citep{yang2020deep}, and DREB \citep{orra2024dynamic}. A detailed description of these baseline trading models is provided in Appendix \ref{baseline}. All learning-based models are trained and evaluated under identical market conditions and transaction cost assumptions to ensure a fair comparison. The proposed BRaG model is evaluated using standard financial performance metrics, including return-based, risk-adjusted, and downside-risk measures. For completeness, the formal definitions of all evaluation metrics are provided in Appendix~\ref{performance}.

\subsection{Results and Discussion}

Table~\ref{tab:results} summarizes the performance of the proposed BRaG framework in comparison with a wide range of baseline strategies across four global stock market indices. The baselines span passive benchmarks, classical optimization methods, technical trading rules, and deep reinforcement learning models, enabling a comprehensive evaluation. Each DRL-based model is trained over five independent runs, and the average values of the evaluation metrics are reported. Due to space constraints, this section reports only the mean results, with the extended results containing both the mean and standard deviation presented in Appendix \ref{extra_results} (Table \ref{tab:results2}). The best-performing results are highlighted in bold.

\begin{table}[!htp]\centering
\caption{Performance comparison of the proposed BRaG model against various baselines across the datasets.}\label{tab:results}
\resizebox{\textwidth}{!}{ 
\begin{tabular}{lccccccccccccccccccc}\toprule
\textbf{Dataset} &\textbf{Metrics} &\textbf{Index} &\textbf{Buy-Hold} &\textbf{MVO} &\textbf{Random} &\textbf{CSMOM} &\textbf{TSMOM} &\textbf{Bollinger} &\textbf{MA Crossover} &\textbf{A2C} &\textbf{DDPG} &\textbf{VS-DRL} &\textbf{SRRS} &\textbf{RSHF} &\textbf{Adaptive} &\textbf{DREB} &\textbf{PPO} &\textbf{$\text{BRaG}^{\dagger}$} \\\midrule
\multirow{5}{*}{\textbf{DJI}} &\textbf{Cumulative Return} &22.8053 &30.561 &19.649 &18.678 &24.241 &25.315 &25.625 &31.926 &37.192 &38.268 &36.959 &12.275 &35.708 &24.023 &42.522 &35.594 &\textbf{53.977} \\
&\textbf{Annual Return} &12.576 &16.522 &10.899 &10.379 &13.334 &13.897 &14.064 &17.323 &20.003 &20.545 &19.885 &6.905 &19.252 &13.219 &22.626 &17.669 &\textbf{28.262} \\
&\textbf{Sharpe Ratio} &0.883 &1.117 &0.976 &0.713 &0.661 &0.914 &0.814 &1.062 &1.067 &1.165 &1.193 &0.457 &1.289 &0.947 &1.405 &1.164 &\textbf{1.472} \\
&\textbf{Max Drawdown} &16.369 &15.916 &\textbf{10.663} &17.142 &29.982 &13.501 &15.991 &18.443 &23.631 &19.724 &19.351 &21.524 &15.701 &16.401 &17.968 &16.208 &17.958 \\
&\textbf{Win Ratio} &54.13 &55.15 &53.44 &47.91 &53.21 &49.31 &54.36 &47.94 &55.05 &54.19 &57.14 &50.01 &54.59 &55.28 &57.34 &55.73 &\textbf{58.05} \\ \midrule
\multirow{5}{*}{\textbf{FTSE}} &\textbf{Cumulative Return} &20.443 &8.604 &23.597 &-14.446 &17.165 &32.061 &27.605 &30.802 &33.917 &29.585 &17.478 &19.144 &33.788 &30.541 &43.898 &32.294 &\textbf{64.241} \\
&\textbf{Annual Return} &11.862 &4.818 &12.891 &-8.523 &9.452 &17.181 &14.914 &16.544 &18.117 &15.923 &9.619 &10.502 &18.052 &16.415 &23.103 &17.298 &\textbf{32.695} \\
&\textbf{Sharpe Ratio} &1.018 &0.426 &1.217 &-0.314 &0.619 &1.232 &0.829 &0.891 &1.034 &1.257 &0.585 &0.709 &1.112 &1.046 &1.376 &1.106 &\textbf{1.632} \\
&\textbf{Max Drawdown} &13.434 &17.293 &\textbf{9.438} &28.404 &16.046 &11.006 &18.551 &17.848 &14.475 &10.004 &16.297 &14.907 &11.817 &14.819 &12.413 &13.458 &11.356 \\
&\textbf{Win Ratio} &55.33 &51.03 &55.13 &26.24 &47.61 &49.66 &49.43 &47.38 &54.42 &51.7 &52.38 &54.21 &53.97 &53.06 &53.74 &54.22 &\textbf{57.51} \\ \midrule
\multirow{5}{*}{\textbf{Sensex}} &\textbf{Cumulative Return} &11.784 &16.135 &11.098 &-14.708 &16.994 &19.191 &12.756 &12.027 &6.008 &5.134 &10.113 &7.423 &23.226 &16.217 &18.669 &21.393 &\textbf{34.067} \\
&\textbf{Annual Return} &6.698 &9.092 &6.316 &-8.842 &9.565 &10.757 &7.237 &3.833 &3.454 &2.956 &5.767 &4.256 &12.924 &9.148 &10.475 &11.943 &\textbf{18.608} \\
&\textbf{Sharpe Ratio} &0.549 &0.698 &0.589 &-0.533 &0.561 &0.604 &0.536 &0.536 &0.303 &0.263 &0.423 &0.325 &0.848 &0.627 &0.664 &0.784 &\textbf{1.042} \\
&\textbf{Max Drawdown} &14.964 &15.913 &17.747 &27.027 &17.935 &21.311 &15.451 &\textbf{11.971} &17.716 &17.833 &23.755 &15.854 &16.235 &15.807 &17.412 &15.354 &15.193 \\
&\textbf{Win Ratio} &51.35 &53.24 &51.19 &27.69 &47.45 &50.46 &48.38 &43.52 &51.16 &51.16 &53.47 &49.39 &53.47 &52.55 &51.39 &51.62 &\textbf{55.07} \\ \midrule
\multirow{5}{*}{\textbf{TWII}} &\textbf{Cumulative Return} &43.276 &51.428 &25.039 &31.014 &62.625 &47.187 &31.188 &17.875 &68.504 &95.384 &29.394 &46.248 &71.962 &49.536 &76.954 &66.454 &\textbf{106.84} \\
&\textbf{Annual Return} &23.954 &28.118 &14.275 &17.506 &33.632 &25.963 &17.968 &10.318 &36.559 &49.478 &16.635 &25.483 &38.226 &27.161 &40.608 &35.566 &\textbf{54.343} \\
&\textbf{Sharpe Ratio} &1.041 &1.053 &1.125 &0.754 &1.072 &1.565 &0.716 &0.671 &1.436 &1.605 &0.747 &0.983 &1.323 &1.213 &1.501 &1.436 &\textbf{1.732} \\
&\textbf{Max Drawdown} &28.693 &33.761 &11.998 &36.567 &38.745 &\textbf{10.762} &38.066 &18.014 &29.551 &26.454 &27.475 &33.783 &30.309 &26.326 &21.789 &26.493 &28.362 \\
&\textbf{Win Ratio} &55.06 &53.68 &57.01 &32.02 &51.31 &52.02 &49.41 &44.42 &57.24 &56.53 &51.78 &53.92 &53.44 &56.06 &54.87 &55.11 &\textbf{60.82} \\
\bottomrule
\end{tabular}}
\end{table}

A clear pattern emerges across all datasets. BRaG achieves the strongest overall performance in terms of cumulative and annualized returns on each market considered. On the DJI dataset, the proposed method delivers a markedly higher final portfolio value than both traditional benchmarks and learning-based strategies. Similar behavior is observed across the FTSE, Sensex, and TWII datasets, where BRaG consistently outperforms the strongest competing methods. The fact that this improvement is observed across markets with different levels of volatility and structural characteristics suggests that the learned policy captures general trading principles rather than dataset-specific effects.

Risk-adjusted performance further distinguishes BRaG from competing methods. The proposed model achieves the highest Sharpe ratio across all datasets, indicating a favorable balance between return and volatility. In contrast, several DRL baselines exhibit higher variability in returns despite achieving competitive profitability. BRaG also maintains comparatively lower maximum drawdowns, highlighting the effectiveness of explicit risk control during action execution. The win ratio analysis indicates that BRaG achieves a higher proportion of profitable trades across all markets. This suggests that performance improvements are driven by consistent decision-making rather than reliance on a small number of extreme outcomes. Technical trading strategies and random trading display lower win ratios and less stable behavior across datasets.

The corresponding out-of-sample cumulative return trajectories are presented in Appendix~\ref{extra_results}, Figure~\ref{fig:plots}. These plots complement the
aggregate results in Table~\ref{tab:results} by showing how the portfolio values evolve over the test period. BRaG exhibits a generally upward trajectory across the four markets, with comparatively controlled declines during several volatile intervals.

Overall, BRaG achieves stronger returns while maintaining a more stable portfolio evolution across different markets. The framework benefits from aggregating diverse expert behavior, which provides a structured starting point for learning, and from adversarial pretraining, which reduces unstable exploration. The additional use of explicit risk control further contributes to limiting drawdowns during periods of market volatility. Additional downside-risk metrics and ablation studies further support these findings and are reported in Appendices~\ref{extra_results} and \ref{ablation}.


\section{Conclusion \label{Conclusion}}
In this paper, we presented BRaG, a framework for stock trading that combines multi-expert imitation learning with adversarial training and explicit risk control. By aggregating heterogeneous expert strategies using a barycenter construction, BRaG learns a stable reference behavior that guides policy learning. Adversarial pretraining helps initialize the policy in a meaningful region of the policy space, which reduces unstable exploration during reinforcement learning. The use of control barrier functions further constrains actions during execution and limits excessive drawdowns. The empirical evaluation shows that BRaG performs well across all the considered markets when compared with both traditional trading rules and recent DRL methods. The framework currently relies on predefined rule-based experts and a simplified risk model. Future work may incorporate richer expert sources, more accurate market dynamics, adaptive risk constraints, and extensions to higher-frequency trading.

\bibliography{iclr2027_conference}

@book{sutton1998reinforcement,
  title={Reinforcement learning: An introduction},
  author={Sutton, Richard S and Barto, Andrew G},
  volume={1},
  number={1},
  year={1998},
  publisher={MIT press Cambridge}
}

@inproceedings{ng2000algorithms,
  title={Algorithms for inverse reinforcement learning.},
  author={Ng, Andrew Y and Russell, Stuart and others},
  booktitle={Icml},
  volume={1},
  number={2},
  pages={2},
  year={2000}
}

@article{fu2017learning,
  title={Learning robust rewards with adversarial inverse reinforcement learning},
  author={Fu, Justin and Luo, Katie and Levine, Sergey},
  journal={arXiv preprint arXiv:1710.11248},
  year={2017}
}

@article{ho2016generative,
  title={Generative adversarial imitation learning},
  author={Ho, Jonathan and Ermon, Stefano},
  journal={Advances in neural information processing systems},
  volume={29},
  year={2016}
}

@article{schulman2017proximal,
  title={Proximal policy optimization algorithms},
  author={Schulman, John and Wolski, Filip and Dhariwal, Prafulla and Radford, Alec and Klimov, Oleg},
  journal={arXiv preprint arXiv:1707.06347},
  year={2017}
}

@article{snoek2012practical,
  title={Practical bayesian optimization of machine learning algorithms},
  author={Snoek, Jasper and Larochelle, Hugo and Adams, Ryan P},
  journal={Advances in neural information processing systems},
  volume={25},
  year={2012}
}

@article{jegadeesh2002cross,
  title={Cross-sectional and time-series determinants of momentum returns},
  author={Jegadeesh, Narasimhan and Titman, Sheridan},
  journal={The Review of Financial Studies},
  volume={15},
  number={1},
  pages={143--157},
  year={2002},
  publisher={Oxford University Press}
}

@article{moskowitz2012time,
  title={Time series momentum},
  author={Moskowitz, Tobias J and Ooi, Yao Hua and Pedersen, Lasse Heje},
  journal={Journal of Financial Economics},
  volume={104},
  number={2},
  pages={228--250},
  year={2012},
  publisher={Elsevier}
}

@article{brock1992simple,
  title={Simple technical trading rules and the stochastic properties of stock returns},
  author={Brock, William and Lakonishok, Josef and LeBaron, Blake},
  journal={The Journal of Finance},
  volume={47},
  number={5},
  pages={1731--1764},
  year={1992},
  publisher={Wiley Online Library}
}

@book{kirkpatrick2010technical,
  title={Technical analysis: the complete resource for financial market technicians},
  author={Kirkpatrick II, Charles D and Dahlquist, Julie R},
  year={2010},
  publisher={FT press}
}

@article{orra2025enhancing,
  title={Enhancing deep reinforcement learning for stock trading: a reward shaping approach via expert feedback},
  author={Orra, Arishi and Choudhary, Himanshu and Sharma, Ankit and Thakur, Manoj},
  journal={Knowledge and Information Systems},
  pages={1--20},
  year={2025},
  publisher={Springer}
}

@inproceedings{orra2024dynamic,
  title={Dynamic reinforced ensemble using bayesian optimization for stock trading},
  author={Orra, Arishi and Bhambu, Aryan and Choudhary, Himanshu and Thakur, Manoj},
  booktitle={Proceedings of the 5th ACM International Conference on AI in Finance},
  pages={361--369},
  year={2024}
}

@inproceedings{yang2020deep,
  title={Deep reinforcement learning for automated stock trading: An ensemble strategy},
  author={Yang, Hongyang and Liu, Xiao-Yang and Zhong, Shan and Walid, Anwar},
  booktitle={Proceedings of the first ACM International Conference on AI in Finance},
  pages={1--8},
  year={2020}
}

@inproceedings{rodinos2023sharpe,
  title={A Sharpe Ratio based reward scheme in Deep Reinforcement Learning for financial trading},
  author={Rodinos, Georgios and Nousi, Paraskevi and Passalis, Nikolaos and Tefas, Anastasios},
  booktitle={IFIP International Conference on Artificial Intelligence Applications and Innovations},
  pages={15--23},
  year={2023},
  organization={Springer}
}

@inproceedings{mnih2016asynchronous,
  title={Asynchronous methods for deep reinforcement learning},
  author={Mnih, Volodymyr and Badia, Adria Puigdomenech and Mirza, Mehdi and Graves, Alex and Lillicrap, Timothy and Harley, Tim and Silver, David and Kavukcuoglu, Koray},
  booktitle={International Conference on Machine Learning},
  pages={1928--1937},
  year={2016},
  organization={PMLR}
}

@article{lillicrap2015continuous,
  title={Continuous control with deep reinforcement learning},
  author={Lillicrap, Timothy P and Hunt, Jonathan J and Pritzel, Alexander and Heess, Nicolas and Erez, Tom and Tassa, Yuval and Silver, David and Wierstra, Daan},
  journal={arXiv preprint arXiv:1509.02971},
  year={2015}
}

@article{zhang2019deep,
  title={Deep reinforcement learning for trading},
  author={Zhang, Zihao and Zohren, Stefan and Roberts, Stephen},
  journal={arXiv preprint arXiv:1911.10107},
  year={2019}
}

@article{markowitz1990foundations,
  title={Foundations of portfolio theory},
  author={Markowitz, Harry M},
  journal={Harry Markowitz: Selected Works, S},
  pages={481--490},
  year={1990},
  publisher={World Scientific}
}

@article{tsantekidis2020price,
  title={Price trailing for financial trading using deep reinforcement learning},
  author={Tsantekidis, Avraam and Passalis, Nikolaos and Toufa, Anastasia-Sotiria and Saitas-Zarkias, Konstantinos and Chairistanidis, Stergios and Tefas, Anastasios},
  journal={IEEE Transactions on neural networks and learning systems},
  volume={32},
  number={7},
  pages={2837--2846},
  year={2020},
  publisher={IEEE}
}

@article{wu2020adaptive,
  title={Adaptive stock trading strategies with deep reinforcement learning methods},
  author={Wu, Xing and Chen, Haolei and Wang, Jianjia and Troiano, Luigi and Loia, Vincenzo and Fujita, Hamido},
  journal={Information Sciences},
  volume={538},
  pages={142--158},
  year={2020},
  publisher={Elsevier}
}

@inproceedings{sun2022deepscalper,
  title={DeepScalper: A risk-aware reinforcement learning framework to capture fleeting intraday trading opportunities},
  author={Sun, Shuo and Xue, Wanqi and Wang, Rundong and He, Xu and Zhu, Junlei and Li, Jian and An, Bo},
  booktitle={Proceedings of the 31st ACM International Conference on Information \& Knowledge Management},
  pages={1858--1867},
  year={2022}
}

@article{kabbani2022deep,
  title={Deep reinforcement learning approach for trading automation in the stock market},
  author={Kabbani, Taylan and Duman, Ekrem},
  journal={IEEE Access},
  volume={10},
  pages={93564--93574},
  year={2022},
  publisher={IEEE}
}

@article{moody1998performance,
  title={Performance functions and reinforcement learning for trading systems and portfolios},
  author={Moody, John and Wu, Lizhong and Liao, Yuansong and Saffell, Matthew},
  journal={Journal of forecasting},
  volume={17},
  number={5-6},
  pages={441--470},
  year={1998},
  publisher={Wiley Online Library}
}

@article{sun2023transaction,
  title={Transaction-aware inverse reinforcement learning for trading in stock markets},
  author={Sun, Qizhou and Gong, Xueyuan and Si, Yain-Whar},
  journal={Applied Intelligence},
  volume={53},
  number={23},
  pages={28186--28206},
  year={2023},
  publisher={Springer Nature BV}
}

@article{roa2019towards,
  title={Towards inverse reinforcement learning for limit order book dynamics},
  author={Roa-Vicens, Jacobo and Chtourou, Cyrine and Filos, Angelos and Rullan, Francisco and Gal, Yarin and Silva, Ricardo},
  journal={arXiv preprint arXiv:1906.04813},
  year={2019}
}

@inproceedings{liu2020adaptive,
  title={Adaptive quantitative trading: An imitative deep reinforcement learning approach},
  author={Liu, Yang and Liu, Qi and Zhao, Hongke and Pan, Zhen and Liu, Chuanren},
  booktitle={Proceedings of the AAAI conference on artificial intelligence},
  volume={34},
  number={02},
  pages={2128--2135},
  year={2020}
}

@article{halperin2022combining,
  title={Combining reinforcement learning and inverse reinforcement learning for asset allocation recommendations},
  author={Halperin, Igor and Liu, Jiayu and Zhang, Xiao},
  journal={arXiv preprint arXiv:2201.01874},
  year={2022}
}

@article{zhang2022reinforcement,
  title={Reinforcement learning for stock prediction and high-frequency trading with T+ 1 rules},
  author={Zhang, Weipeng and Yin, Tao and Zhao, Yunan and Han, Bing and Liu, Huanxi},
  journal={IEEE Access},
  volume={11},
  pages={14115--14127},
  year={2022},
  publisher={IEEE}
}

@article{choudhary2025risk,
  title={Risk-adjusted deep reinforcement learning for portfolio optimization: A multi-reward approach},
  author={Choudhary, Himanshu and Orra, Arishi and Sahoo, Kartik and Thakur, Manoj},
  journal={International Journal of Computational Intelligence Systems},
  volume={18},
  number={1},
  pages={126},
  year={2025},
  publisher={Springer}
}

@article{choudhary2026cvar,
  title={A CVaR-Constrained Safe Reinforcement Learning Framework with Action Repair for Practical Portfolio Optimization},
  author={Choudhary, Himanshu and Orra, Arishi and Thakur, Manoj and Gao, Xiao-Zhi and Sahu, Prabhat Kumar},
  journal={IEEE Transactions on Artificial Intelligence},
  year={2026},
  publisher={IEEE}
}

@article{choudhary2026dynamic,
  title={Dynamic Asset Pre-selection Guided Portfolio Optimization Using Deep Reinforcement Learning},
  author={Choudhary, Himanshu and Orra, Arishi and Thakur, Manoj},
  journal={Annals of Data Science},
  pages={1--45},
  year={2026},
  publisher={Springer}
}
\bibliographystyle{iclr2027_conference}

\appendix
\section{Proximal Policy Optimization \label{ppo}}

Proximal Policy Optimization (PPO) is a first-order policy gradient method designed to provide stable and efficient policy updates in reinforcement learning. PPO belongs to the family of actor-critic algorithms and optimizes a parameterized policy \(\pi_\theta(a|s)\) using stochastic gradient ascent while constraining the update to remain close to the previous policy \(\pi_{\theta_{\text{old}}}\). This constraint mitigates the instability often observed in vanilla policy gradient methods when large updates are applied. PPO achieves this objective without explicitly solving a constrained optimization problem, which makes it computationally efficient and well-suited for large-scale applications.

The core idea of PPO is to maximize a clipped surrogate objective.  
Let
\[
r_t(\theta)=\frac{\pi_\theta(a_t|s_t)}{\pi_{\theta_{\text{old}}}(a_t|s_t)}
\]
denote the probability ratio between the new and old policies, and let \(\hat A_t\) be an estimate of the advantage function.  
The PPO objective is defined as
\begin{equation}
L^{\text{PPO}}(\theta)=
\mathbb{E}\Big[
\min\big(
r_t(\theta)\hat A_t,\;
\text{clip}(r_t(\theta),1-\epsilon,1+\epsilon)\hat A_t
\big)
\Big],
\end{equation}
where \(\epsilon > 0\) is a hyperparameter that controls the size of the trust region.  
This clipping operation prevents excessively large policy updates by limiting the contribution of samples where the new policy deviates significantly from the old one \citep{schulman2017proximal}. In practice, PPO is combined with a learned value function and entropy regularization to reduce variance and encourage exploration. 

\section{Expert Trajectories \label{expert}}

To construct diverse and informative expert demonstrations, we employ a set of rule-based technical trading strategies that have been well-studied in the finance literature. These strategies capture distinct market behaviors and provide complementary perspectives on trend persistence and price deviations. 

\begin{enumerate}
    \item \textbf{Time-Series Momentum (TSMOM):} TSMOM is a momentum-based strategy that takes positions based on the sign of an asset's own past returns. For an asset $i$ at time $t$, the trading signal is given by
    \[
    s_{i,t} = \mathrm{sign}\big(r_{i,t-L:t-1}\big),
    \]
    where $r_{i,t-L:t-1}$ denotes the cumulative return over a lookback window of length $L$. A long position is taken when the signal is positive, and a short position otherwise. TSMOM belongs to the class of time-series momentum strategies. It is beneficial as it captures persistent trends within individual assets and adapts naturally to regime changes where trends strengthen or weaken over time \citep{moskowitz2012time}.

    \item \textbf{Cross-Sectional Momentum (CSMOM):} CSMOM ranks assets based on their past returns relative to one another and allocates positions accordingly. Let $r_{i,t-L:t-1}$ denote the past return of asset $i$. Assets are ranked across the universe, and the signal is defined as
    \[
    s_{i,t} = \mathrm{rank}\big(r_{i,t-L:t-1}\big),
    \]
    with long positions assigned to top-ranked assets and short positions to bottom-ranked assets. CSMOM is a momentum-based cross-sectional strategy. It is useful because it exploits relative performance differences across assets and provides diversification benefits through long-short positioning \citep{jegadeesh2002cross}.
    
    \item \textbf{Moving Average Crossover:} The moving average crossover strategy generates signals based on the interaction of short-term and long-term moving averages of price. Let $\text{MA}_f(t)$ and $\text{MA}_s(t)$ denote fast and slow moving averages, respectively. The trading signal is
    \[
    s_t =
    \begin{cases}
    +1, & \text{if } \text{MA}_f(t) > \text{MA}_s(t), \\
    -1, & \text{otherwise}.
    \end{cases}
    \]
    This strategy belongs to the trend-following category. It is beneficial due to its simplicity and its ability to filter out short-term noise while responding to sustained price movements \citep{brock1992simple}.
    
    \item \textbf{Bollinger Bands:} Bollinger Bands are constructed using a moving average and a volatility-based envelope. Let $\mu_t$ and $\sigma_t$ denote the rolling mean and standard deviation of prices. The upper and lower bands are defined as
    \[
    \text{UB}_t = \mu_t + k\sigma_t, \quad \text{LB}_t = \mu_t - k\sigma_t.
    \]
    Trading signals are generated when prices deviate significantly from these bands. This strategy is commonly associated with mean-reversion behavior. It is useful as it captures overbought and oversold conditions, and complements momentum-based experts by focusing on price corrections rather than trend continuation \citep{kirkpatrick2010technical}.

\end{enumerate}

\section{Baseline Trading Strategies \label{baseline}} 

The performance of the proposed BRaG framework is benchmarked against a broad spectrum of trading strategies to ensure a comprehensive evaluation. These baselines include:

\paragraph{Rule-Based Technical Experts} 
The individual performance of the CSMOM, TSMOM, MA Crossover, and Bollinger Band strategies.

\paragraph{Standard DRL Agents}
Vanilla A2C \citep{mnih2016asynchronous}, DDPG \citep{lillicrap2015continuous}, and PPO \citep{schulman2017proximal} algorithms.

\paragraph{Traditional Finance Baselines}
\begin{itemize}
    \item \textbf{Market Index:} This represents a hypothetical portfolio used to measure the broader macroeconomic performance. Comparing against the index reveals whether the active trading strategy genuinely generated excess alpha over the general market drift.
    
    \item \textbf{Buy-and-Hold:} The buy-and-hold strategy involves buying and holding the assets over the entire investment horizon without active trading. It assumes no rebalancing, and benefits from long-term market appreciation.

    \item \textbf{Random Trading:} Random trading is a naive baseline where trades are executed randomly without any predictive model or signal. It functions as a statistical floor for performance.

    \item \textbf{ Mean-Variance Optimization (MVO):} Mean-Variance Optimization is a foundational portfolio construction technique that aims to balance expected return against risk. It relies on estimates of expected returns and covariances, and assumes normally distributed returns, forming a foundational approach in modern portfolio theory.
\end{itemize}

\paragraph{State-of-the-Art Literature}
\begin{itemize}
    \item \textbf{Adaptive:} An ensemble trading strategy combining three actor-critic algorithms: A2C, DDPG, and PPO, to leverage their complementary strengths \citep{yang2020deep}.

    \item \textbf{Volatility-Scaled Reward (VS-DRL):} This study incorporates volatility scaling into the reward function to dynamically adjust position sizes based on market volatility, thereby enhancing robustness and risk management \citep{zhang2019deep}.

    \item \textbf{Sharpe Ratio Reward Scheme (SRRS):} This work implements a reward shaping technique by directly integrating an approximation of the Sharpe ratio into the reward function alongside profit and loss \citep{rodinos2023sharpe}.

    \item \textbf{Dynamic Reinforced Ensemble (DREB):}  A dynamic ensemble method that uses Bayesian optimization to assign time-varying weights to multiple DRL agents for automated stock trading \citep{orra2024dynamic}.

    \item \textbf{Reward Shaping via Human Feedback (RSHF):} A reward-shaping framework that augments the conventional profit-and-loss objective with trading rules derived from technical indicators as human feedback to improve trading accuracy \citep{orra2025enhancing}.
\end{itemize}

\section{Performance Metrics \label{performance}}

We evaluate all trading strategies using a set of standard financial performance metrics that capture profitability, risk-adjusted return, and downside risk.

    \begin{enumerate}
        \item \textbf{Cumulative Return:} The net change in portfolio value between the beginning and end of the trading period.
        
        \item \textbf{Annualized Return:} It represents the average yearly growth rate of the portfolio.
        
        \item \textbf{Sharpe Ratio:} It evaluates risk-adjusted performance by comparing the mean excess return of the strategy to the standard deviation of returns, thereby quantifying return per unit of risk.
        
        \item \textbf{Maximum Drawdown:} Denotes the most severe loss experienced by the portfolio during the trading period, measured as the largest observed decline in portfolio value from a historical peak to a subsequent trough.
        
        \item \textbf{Win Ratio:} The percentage of executed trades that yield a positive outcome compared to the total number of trades.
    \end{enumerate}

\section{Hyperparameter Setting \label{hyperparameter}}

All experiments are conducted under uniform training and evaluation settings to ensure fair comparison across models. Hyperparameters for the proposed BRaG framework are selected using Bayesian optimization, which provides an efficient search strategy in high-dimensional parameter spaces. The tuning process is performed on a held-out validation split derived from the training data. We employ the Hyperopt library, which implements tree-structured Parzen estimators to guide the optimization. The search ranges for policy, discriminator, and optimization-related hyperparameters are chosen based on established empirical practices and preliminary experiments. A complete list of tuned hyperparameters and their corresponding ranges is reported in Table~\ref{tab:hyperparameter}.

\begin{table}[!htp]\centering
\caption{Hyperparameter configuration and search space for the proposed BRaG model.}\label{tab:hyperparameter}
\begin{tabular}{lccc}\toprule
\textbf{Model } &\textbf{Hyperparameter} &\textbf{Range} \\\midrule
\multirow{9}{*}{PPO} &Hidden Dimension &[2,512] \\
&Number of Layers &[1,8] \\
&Activation Function &[ReLU, Tanh, Sigmoid] \\
&Learning Rate & $[e^{-8}, e^{-1}]$ \\
&Dropout &[0,0.5] \\
&$\gamma$ &[0.9,0.99] \\
&PPO Epochs &[5,50] \\
&Value Coefficient &[0.01,0.5] \\
&Entropy Coefficient &[0.01,0.1] \\ \midrule
\multirow{4}{*}{GAIL} &Hidden Dimension &[2,512] \\
&Number of Layers &[1,8] \\
&Activation Function &[ReLU, Tanh, Sigmoid] \\
&Dropout &[0,0.5] \\ \midrule
\multirow{4}{*}{CBF} &Rolling Window &20 \\
&$\kappa$ &1 \\
&$\alpha$ &0.1 \\
&CBF penalty coefficient &[0.01,0.1] \\
\bottomrule
\end{tabular}
\end{table}

\section{Additional Results \label{extra_results}}

    \begin{figure}[!htp]
        \centering
        \begin{subfigure}[t]{0.48\textwidth}
            \centering
            \includegraphics[width=\linewidth, height=0.18\textheight,keepaspectratio]{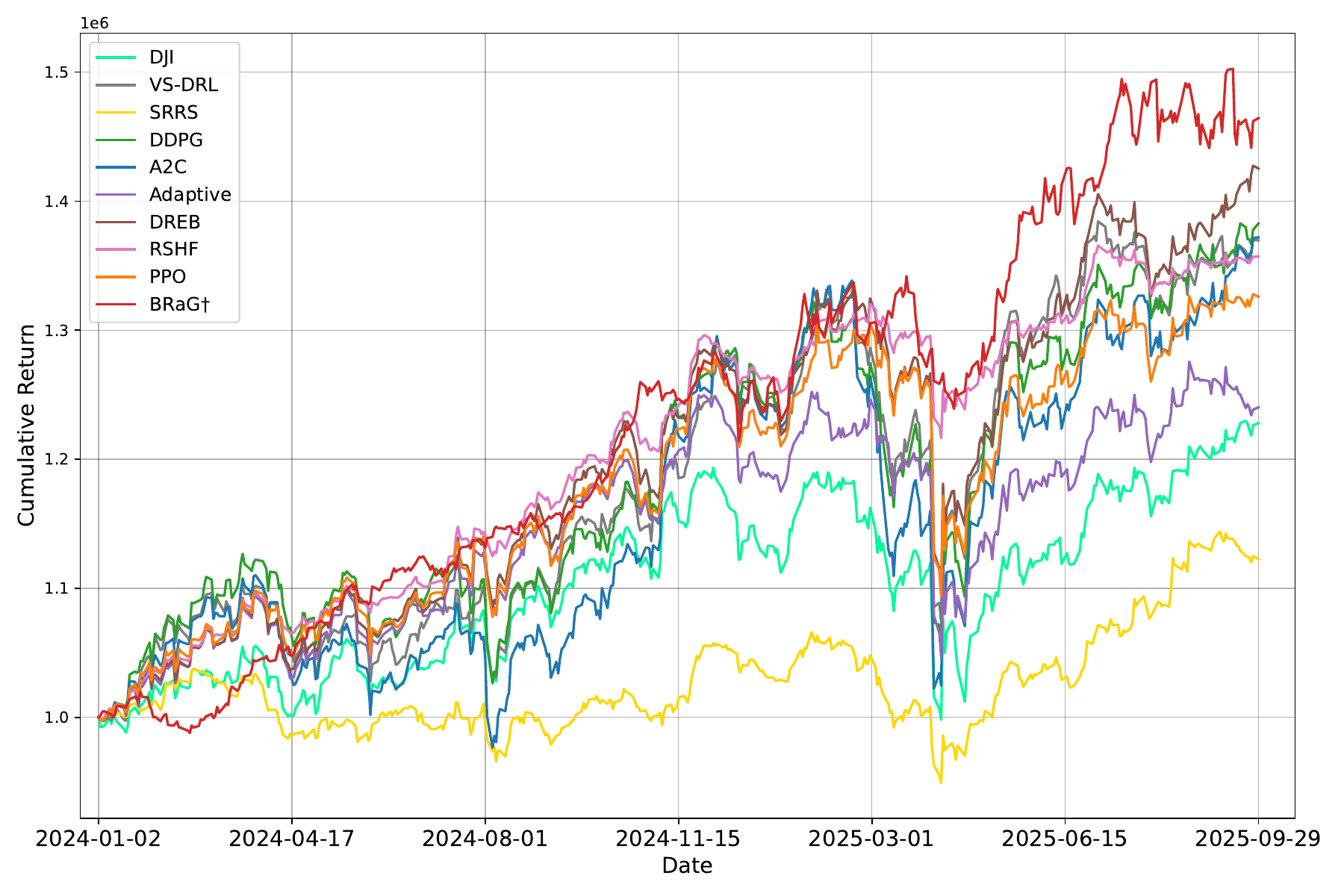}
            \caption{DJI}
            \label{fig:dataset_metric1}
        \end{subfigure}
        \hfill
        \begin{subfigure}[t]{0.48\textwidth}
            \centering
            \includegraphics[width=\linewidth, height=0.18\textheight,keepaspectratio]{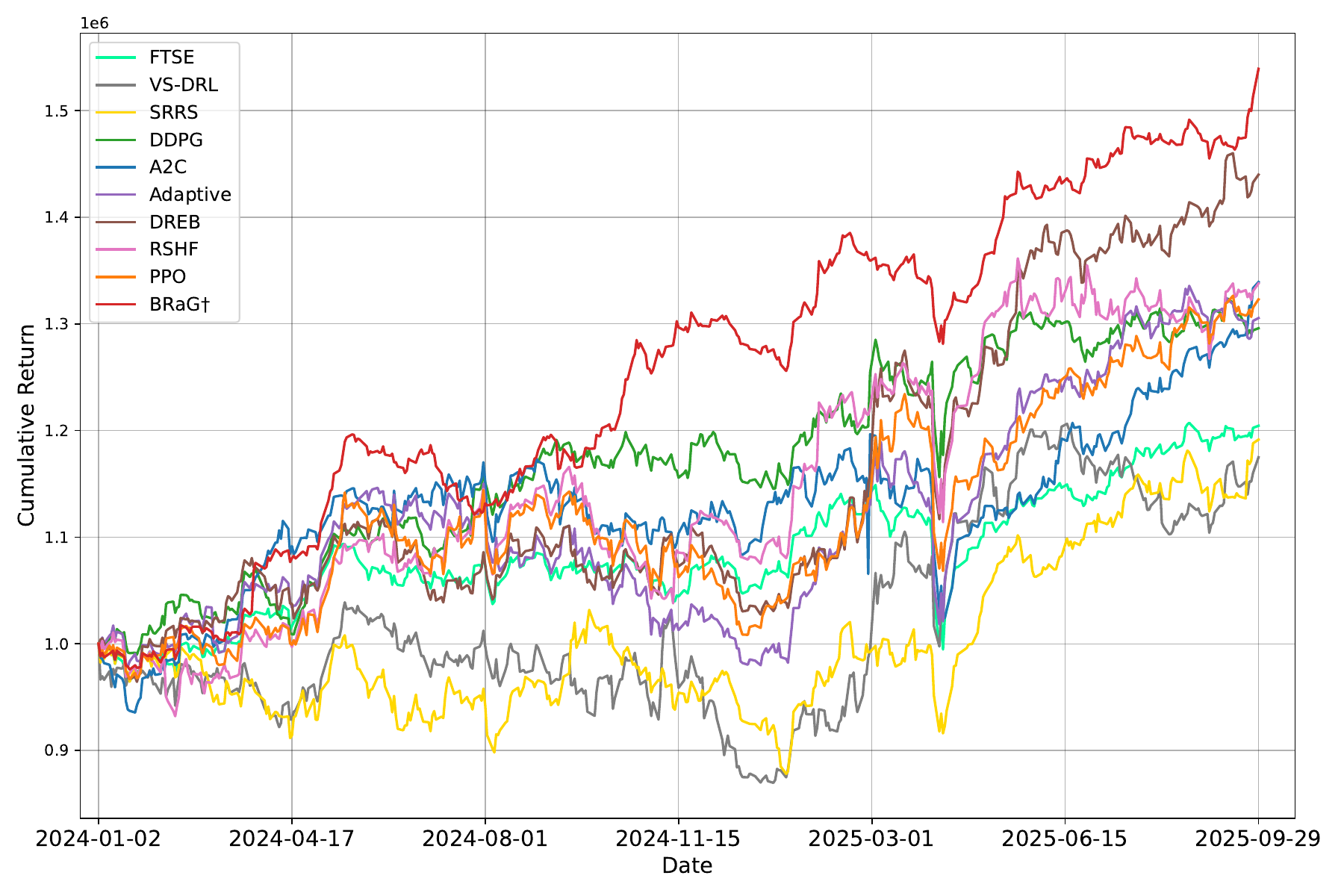}
            \caption{FTSE}
            \label{fig:dataset_metric2}
        \end{subfigure}
    
    
        \begin{subfigure}[t]{0.48\textwidth}
            \centering
            \includegraphics[width=\linewidth, height=0.18\textheight,keepaspectratio]{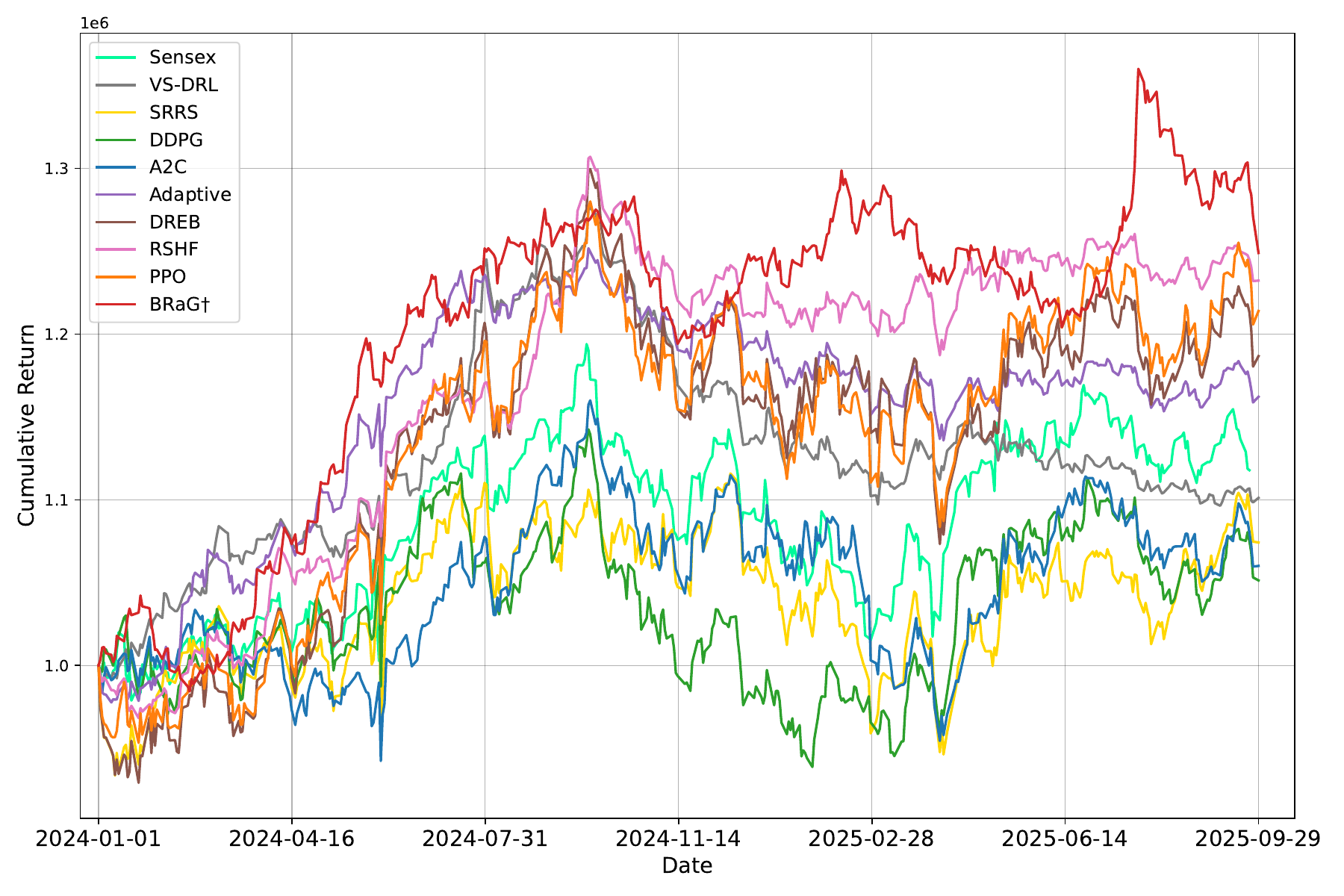}
            \caption{Sensex}
            \label{fig:dataset_metric3}
        \end{subfigure}
        \hfill
        \begin{subfigure}[t]{0.48\textwidth}
            \centering
            \includegraphics[width=\linewidth, height=0.18\textheight,keepaspectratio]{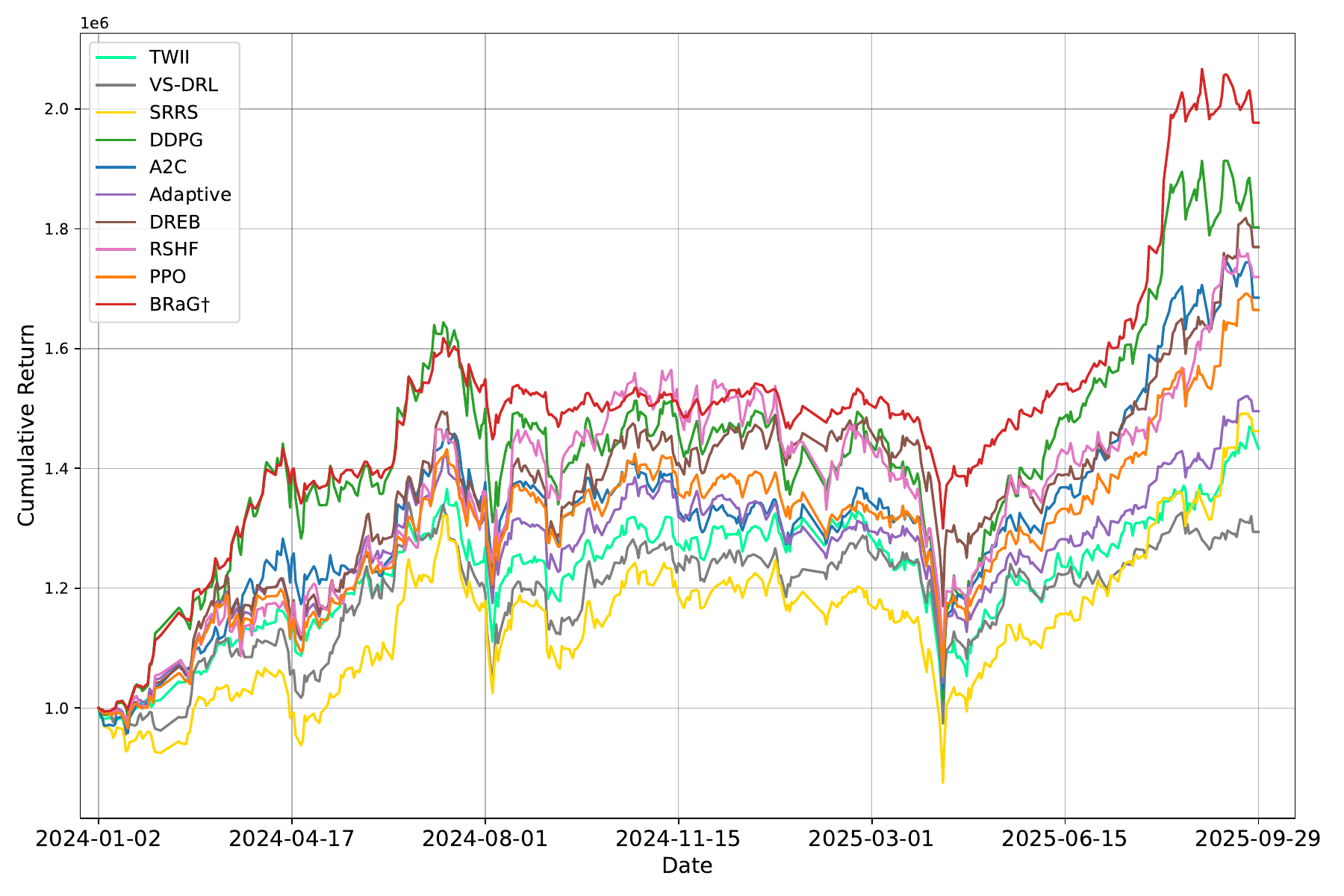}
            \caption{TWII}
            \label{fig:dataset_metric4}
        \end{subfigure}
    
        \caption{Cumulative return trajectories of the proposed BRaG model and baseline strategies across the DJI, FTSE, Sensex, and TWII datasets during the out-of-sample trading period.}
        \label{fig:plots}
    \end{figure}

The cumulative return plots in Figure~\ref{fig:plots} illustrate how the trading strategies behave over time rather than only at the end of the evaluation period. Across all four markets, BRaG shows a relatively steady upward trajectory, with fewer abrupt declines compared to most learning-based baselines. In the DJI and FTSE plots, several DRL methods achieve short phases of rapid growth but suffer noticeable reversals during volatile intervals. BRaG, in contrast, appears to moderate its exposure during these periods and recovers more gradually after downturns. A similar pattern is visible in the Sensex and TWII plots, where technical strategies and some DRL agents display uneven growth and larger fluctuations. Although certain rule-based methods perform competitively in specific sub-periods, their performance is not consistent across the whole horizon. Random trading and passive strategies remain clearly dominated in all markets. The smoother equity curves observed for BRaG suggest that the policy avoids extreme positions and benefits from a more conservative response to unfavorable market movements, which is consistent with the use of expert-guided pretraining and explicit risk constraints.

Table \ref{tab:results2} presents a comprehensive performance comparison, including the mean and standard deviation across five independent random seeds, for the learning-based methods (deterministic baselines inherently exhibit zero variance). The expanded results further underscore the robustness of the proposed framework. While standard DRL models, such as A2C, DDPG, and PPO, demonstrate moderate to high performance variance across different initializations, BRaG not only achieves the highest mean performance across all datasets but also maintains notably tighter standard deviations. This reduction in variance demonstrates stable behavior across different random initializations.

\begin{table}[!htp]\centering
\caption{ Performance comparison of the proposed BRaG model against various baselines across the datasets. Results are reported as mean $\pm$ standard deviation over five independent runs.} \label{tab:results2}
\resizebox{\textwidth}{!}{ 
\begin{tabular}{lcccccccccccc}\toprule
\textbf{Dataset} &\textbf{Metrics} &\textbf{Random} &\textbf{A2C} &\textbf{DDPG} &\textbf{VS-DRL} &\textbf{SRRS} &\textbf{RSHF} &\textbf{Adaptive} &\textbf{DREB} &\textbf{PPO} &\textbf{$\text{BRaG}^{\dagger}$} \\\midrule
\multirow{5}{*}{\textbf{DJI}} &\textbf{Cumulative Return} &18.68 $\pm$ 5.42 &37.19 $\pm$ 3.41 &38.27 $\pm$ 4.12 &36.96 $\pm$ 3.85 &12.28 $\pm$ 2.15 &35.71 $\pm$ 2.94 &24.02 $\pm$ 2.65 &42.52 $\pm$ 3.10 &35.59 $\pm$ 2.87 &\textbf{53.98 $\pm$ 1.95} \\
&\textbf{Annual Return} &10.38 $\pm$ 3.15 &20.00 $\pm$ 1.84 &20.55 $\pm$ 2.21 &19.89 $\pm$ 1.95 &6.91 $\pm$ 1.12 &19.25 $\pm$ 1.65 &13.22 $\pm$ 1.43 &22.63 $\pm$ 1.75 &17.67 $\pm$ 1.52 &\textbf{28.26 $\pm$ 1.05} \\
&\textbf{Sharpe Ratio} &0.71 $\pm$ 0.18 &1.07 $\pm$ 0.12 &1.17 $\pm$ 0.14 &1.19 $\pm$ 0.13 &0.46 $\pm$ 0.16 &1.29 $\pm$ 0.11 &0.95 $\pm$ 0.14 &1.41 $\pm$ 0.09 &1.16 $\pm$ 0.10 &\textbf{1.47 $\pm$ 0.06} \\
&\textbf{Max Drawdown} &17.14 $\pm$ 3.85 &23.63 $\pm$ 2.15 &19.72 $\pm$ 2.84 &19.35 $\pm$ 2.45 &21.52 $\pm$ 3.12 &\textbf{15.70 $\pm$ 1.85} &16.40 $\pm$ 2.05 &17.97 $\pm$ 1.75 &16.21 $\pm$ 1.64 &17.96 $\pm$ 1.25 \\
&\textbf{Win Ratio} &47.91 $\pm$ 2.45 &55.05 $\pm$ 1.34 &54.19 $\pm$ 1.65 &57.14 $\pm$ 1.45 &50.01 $\pm$ 1.85 &54.59 $\pm$ 1.25 &55.28 $\pm$ 1.55 &57.34 $\pm$ 1.15 &55.73 $\pm$ 1.15 &\textbf{58.05 $\pm$ 0.85} \\\midrule
\multirow{5}{*}{\textbf{FTSE}} &\textbf{Cumulative Return} &-14.45 $\pm$ 6.15 &33.92 $\pm$ 3.85 &29.59 $\pm$ 4.15 &17.48 $\pm$ 3.25 &19.14 $\pm$ 3.65 &33.79 $\pm$ 3.15 &30.54 $\pm$ 2.95 &43.90 $\pm$ 3.25 &32.29 $\pm$ 3.14 &\textbf{64.24 $\pm$ 2.41} \\
&\textbf{Annual Return} &-8.52 $\pm$ 4.12 &18.12 $\pm$ 2.15 &15.92 $\pm$ 2.45 &9.62 $\pm$ 1.85 &10.50 $\pm$ 1.95 &18.05 $\pm$ 1.75 &16.42 $\pm$ 1.65 &23.10 $\pm$ 1.85 &17.30 $\pm$ 1.75 &\textbf{32.70 $\pm$ 1.25} \\
&\textbf{Sharpe Ratio} &-0.31 $\pm$ 0.22 &1.03 $\pm$ 0.14 &1.26 $\pm$ 0.16 &0.59 $\pm$ 0.18 &0.71 $\pm$ 0.15 &1.11 $\pm$ 0.12 &1.05 $\pm$ 0.14 &1.38 $\pm$ 0.10 &1.11 $\pm$ 0.11 &\textbf{1.63 $\pm$ 0.07} \\
&\textbf{Max Drawdown} &28.40 $\pm$ 4.25 &14.48 $\pm$ 2.25 &\textbf{10.00 $\pm$ 1.85} &16.30 $\pm$ 2.45 &14.91 $\pm$ 2.15 &11.82 $\pm$ 1.65 &14.82 $\pm$ 1.95 &12.41 $\pm$ 1.55 &13.46 $\pm$ 1.65 &11.36 $\pm$ 1.15 \\
&\textbf{Win Ratio} &26.24 $\pm$ 3.15 &54.42 $\pm$ 1.45 &51.70 $\pm$ 1.85 &52.38 $\pm$ 1.75 &54.21 $\pm$ 1.65 &53.97 $\pm$ 1.35 &53.06 $\pm$ 1.55 &53.74 $\pm$ 1.25 &54.22 $\pm$ 1.35 &\textbf{57.51 $\pm$ 0.95} \\\midrule
\multirow{5}{*}{\textbf{Sensex}} &\textbf{Cumulative Return} &-14.71 $\pm$ 5.85 &6.01 $\pm$ 2.85 &5.13 $\pm$ 3.15 &10.11 $\pm$ 2.95 &7.42 $\pm$ 2.45 &23.23 $\pm$ 2.75 &16.22 $\pm$ 2.55 &18.67 $\pm$ 2.15 &21.39 $\pm$ 2.75 &\textbf{34.07 $\pm$ 1.85} \\
&\textbf{Annual Return} &-8.84 $\pm$ 3.65 &3.45 $\pm$ 1.65 &2.96 $\pm$ 1.85 &5.77 $\pm$ 1.75 &4.26 $\pm$ 1.45 &12.92 $\pm$ 1.55 &9.15 $\pm$ 1.45 &10.48 $\pm$ 1.35 &11.94 $\pm$ 1.55 &\textbf{18.61 $\pm$ 1.05} \\
&\textbf{Sharpe Ratio} &-0.53 $\pm$ 0.24 &0.30 $\pm$ 0.15 &0.26 $\pm$ 0.18 &0.42 $\pm$ 0.16 &0.33 $\pm$ 0.14 &0.85 $\pm$ 0.12 &0.63 $\pm$ 0.13 &0.66 $\pm$ 0.11 &0.78 $\pm$ 0.12 &\textbf{1.04 $\pm$ 0.08} \\
&\textbf{Max Drawdown} &27.03 $\pm$ 4.15 &17.72 $\pm$ 2.45 &17.83 $\pm$ 2.85 &23.76 $\pm$ 3.15 &15.85 $\pm$ 2.25 &16.24 $\pm$ 1.95 &15.81 $\pm$ 2.15 &17.41 $\pm$ 1.85 &15.35 $\pm$ 1.95 &\textbf{15.19 $\pm$ 1.35} \\
&\textbf{Win Ratio} &27.69 $\pm$ 2.95 &51.16 $\pm$ 1.65 &51.16 $\pm$ 1.95 &53.47 $\pm$ 1.85 &49.39 $\pm$ 1.75 &53.47 $\pm$ 1.45 &52.55 $\pm$ 1.55 &51.39 $\pm$ 1.35 &51.62 $\pm$ 1.45 &\textbf{55.07 $\pm$ 0.95} \\\midrule
\multirow{5}{*}{\textbf{TWII}} &\textbf{Cumulative Return} &31.01 $\pm$ 6.85 &68.50 $\pm$ 4.85 &95.38 $\pm$ 6.15 &29.39 $\pm$ 4.25 &46.25 $\pm$ 4.65 &71.96 $\pm$ 4.15 &49.54 $\pm$ 3.95 &76.95 $\pm$ 4.35 &66.45 $\pm$ 4.12 &\textbf{106.84 $\pm$ 3.55} \\
&\textbf{Annual Return} &17.51 $\pm$ 3.95 &36.56 $\pm$ 2.65 &49.48 $\pm$ 3.45 &16.64 $\pm$ 2.45 &25.48 $\pm$ 2.55 &38.23 $\pm$ 2.35 &27.16 $\pm$ 2.25 &40.61 $\pm$ 2.45 &35.57 $\pm$ 2.25 &\textbf{54.34 $\pm$ 1.95} \\
&\textbf{Sharpe Ratio} &0.75 $\pm$ 0.19 &1.44 $\pm$ 0.13 &1.61 $\pm$ 0.08 &0.75 $\pm$ 0.17 &0.98 $\pm$ 0.15 &1.32 $\pm$ 0.12 &1.21 $\pm$ 0.14 &1.50 $\pm$ 0.11 &1.44 $\pm$ 0.12 &\textbf{1.73 $\pm$ 0.16} \\
&\textbf{Max Drawdown} &36.57 $\pm$ 5.15 &29.55 $\pm$ 3.45 &26.45 $\pm$ 4.15 &27.48 $\pm$ 3.65 &33.78 $\pm$ 3.85 &30.31 $\pm$ 3.15 &26.33 $\pm$ 3.25 &\textbf{21.79 $\pm$ 2.85} &26.49 $\pm$ 2.95 &28.36 $\pm$ 2.15 \\
&\textbf{Win Ratio} &32.02 $\pm$ 3.85 &57.24 $\pm$ 1.85 &56.53 $\pm$ 2.15 &51.78 $\pm$ 1.95 &53.92 $\pm$ 1.85 &53.44 $\pm$ 1.65 &56.06 $\pm$ 1.75 &54.87 $\pm$ 1.55 &55.11 $\pm$ 1.65 &\textbf{60.82 $\pm$ 1.15} \\
\bottomrule
\end{tabular}}
\end{table}

\begin{figure}[!htp]
    \centering
    \begin{subfigure}[t]{0.49\textwidth}
        \centering
        \includegraphics[
            width=\linewidth,
            height=0.50\textheight,
            keepaspectratio
        ]{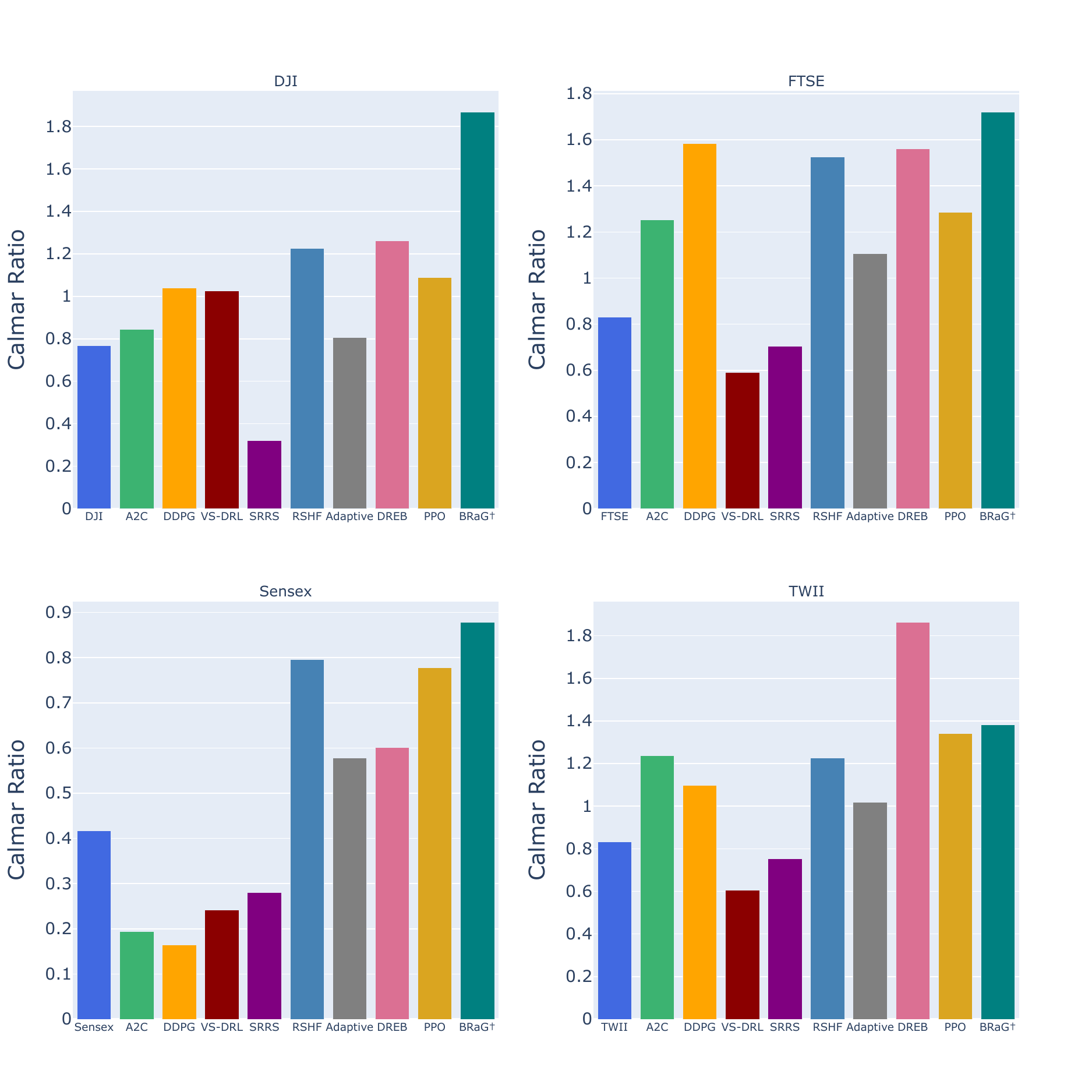}
        \caption{Calmar Ratio}
        \label{fig:calmar}
    \end{subfigure}
    \hfill
    \begin{subfigure}[t]{0.49\textwidth}
        \centering
        \includegraphics[
            width=\linewidth,
            height=0.50\textheight,
            keepaspectratio
        ]{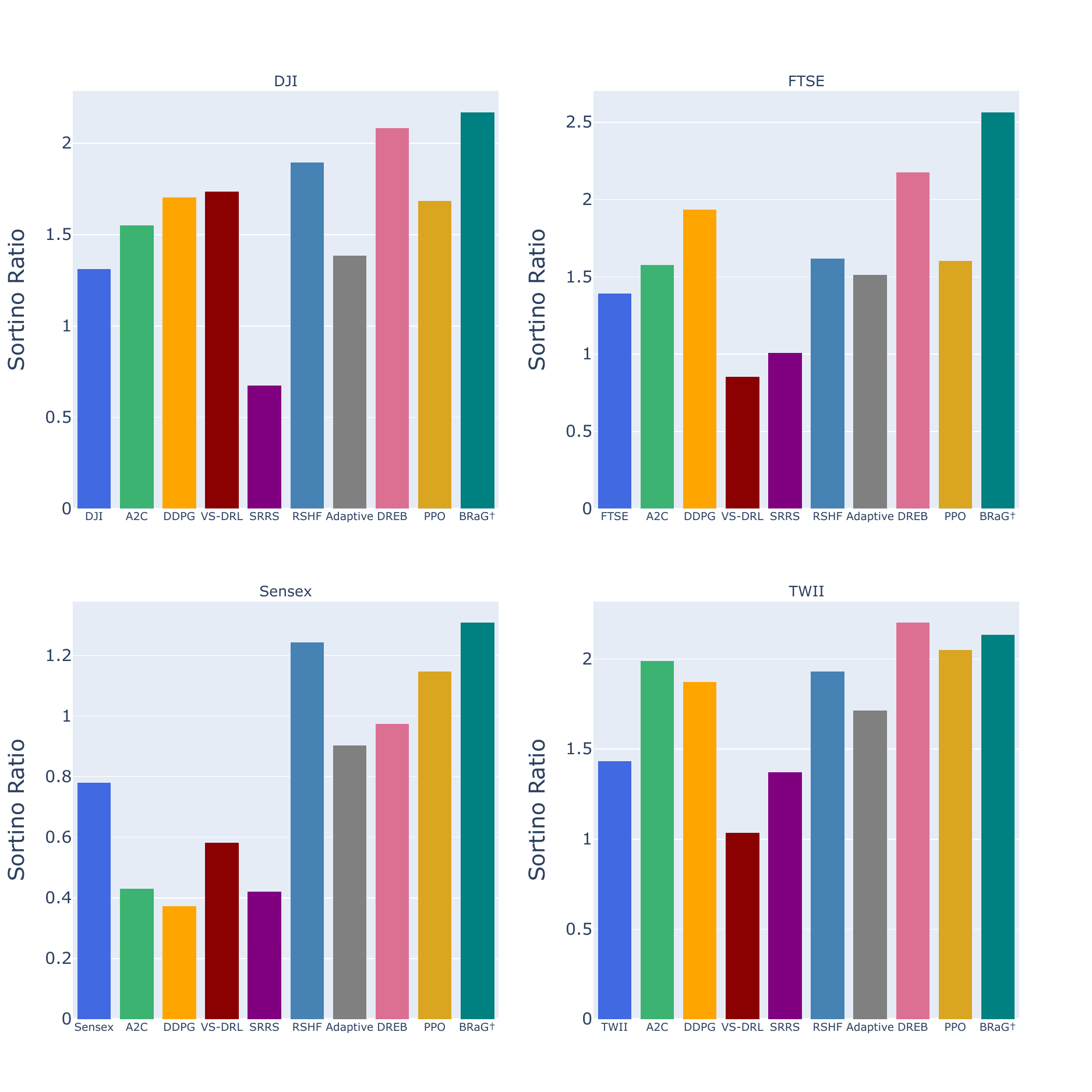}
        \caption{Sortino ratio}
        \label{fig:sortino}
    \end{subfigure}

    \caption{Distribution of Calmar and Sortino ratios across all models and datasets, illustrating comparative downside risk and risk-adjusted performance.}
    \label{fig:additional_plots}
\end{figure}

To further examine the risk characteristics of the proposed method, we report additional results based on the Calmar and Sortino ratios. These metrics are commonly used to assess downside risk and capital preservation, which are not fully captured by return-based measures alone. For this purpose, we plot the distribution of Calmar and Sortino ratios across all models and datasets. The corresponding histograms are shown in Figures~\ref{fig:calmar} and \ref{fig:sortino}. These plots provide a complementary view of model behavior under adverse market movements.

The Calmar ratio compares overall returns with the maximum drawdown experienced during the trading period. BRaG consistently attains higher Calmar values across the DJI, FTSE, Sensex, and TWII datasets. This suggests that the gains achieved by BRaG are not driven by taking excessive drawdown risk. Several DRL baselines achieve reasonable returns but exhibit lower Calmar ratios, indicating larger drawdowns during unfavorable periods. Rule-based strategies and random trading show noticeably weaker performance, particularly on markets with higher volatility. Overall, the Calmar results indicate that BRaG is able to control large losses more effectively than the competing approaches. The Sortino ratio focuses specifically on downside variability and penalizes negative returns rather than total volatility. BRaG again achieves the highest Sortino ratios across all markets. This indicates that negative return fluctuations are comparatively limited for the proposed method. While some learning-based baselines perform well on individual datasets, their Sortino ratios vary substantially across markets. In contrast, BRaG shows more stable behavior, with consistently strong downside-adjusted performance. These observations support the view that combining expert-guided pretraining with explicit risk constraints leads to more controlled trading behavior.

\section{Ablation Study \label{ablation}}

\begin{figure}[!htp]
  \begin{center}
    \centerline{\includegraphics[width=\columnwidth, height=6cm, keepaspectratio]{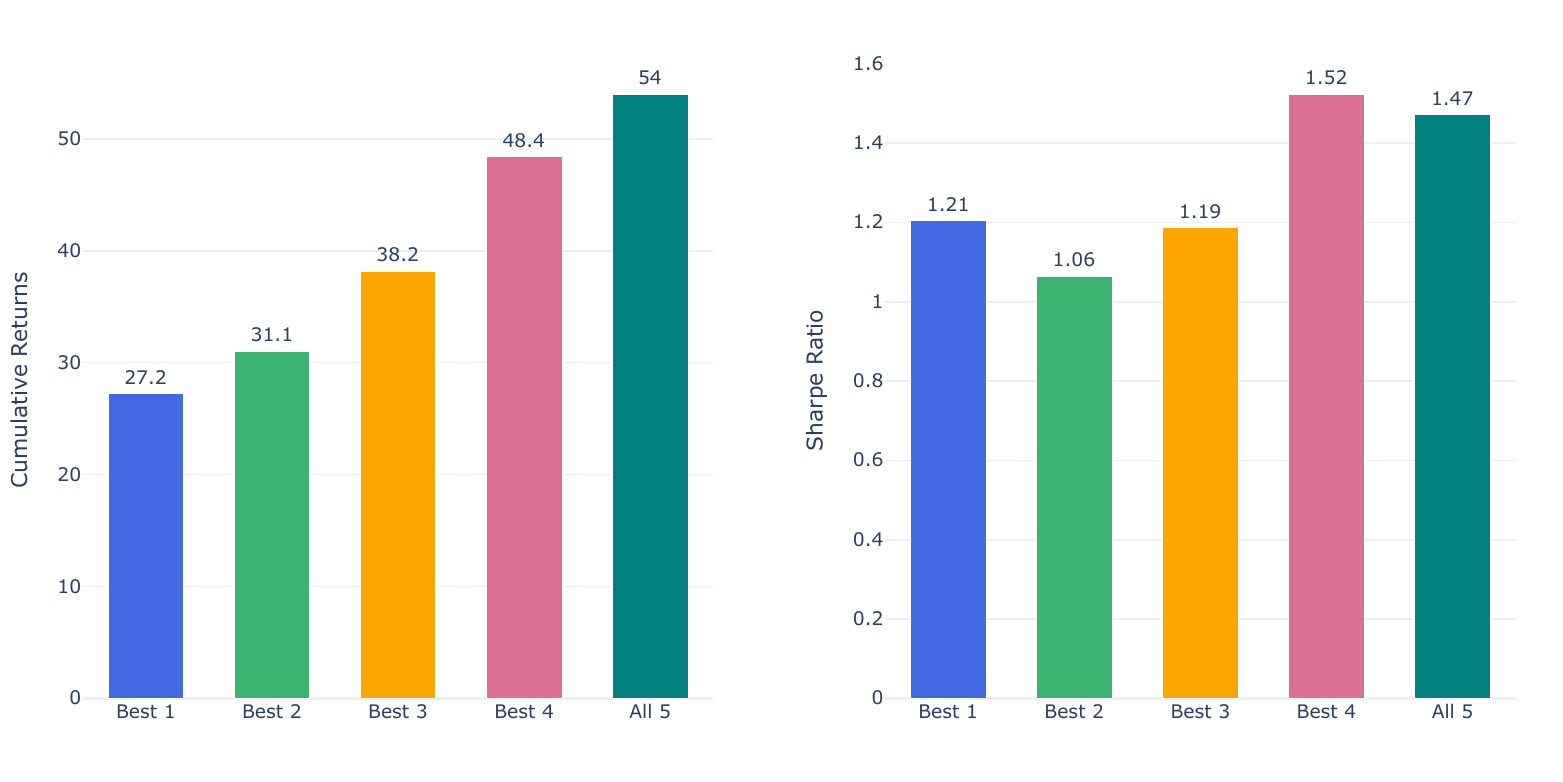}}
    \caption{Effect of varying the number of expert trajectories used for barycenter construction on cumulative return and Sharpe ratio. Experts are added incrementally based on validation performance.}
    \label{fig:expert_ablation}
  \end{center}
\end{figure}

\begin{figure}[!htp]
  \begin{center}
    \centerline{\includegraphics[width=\columnwidth, height=6cm, keepaspectratio]{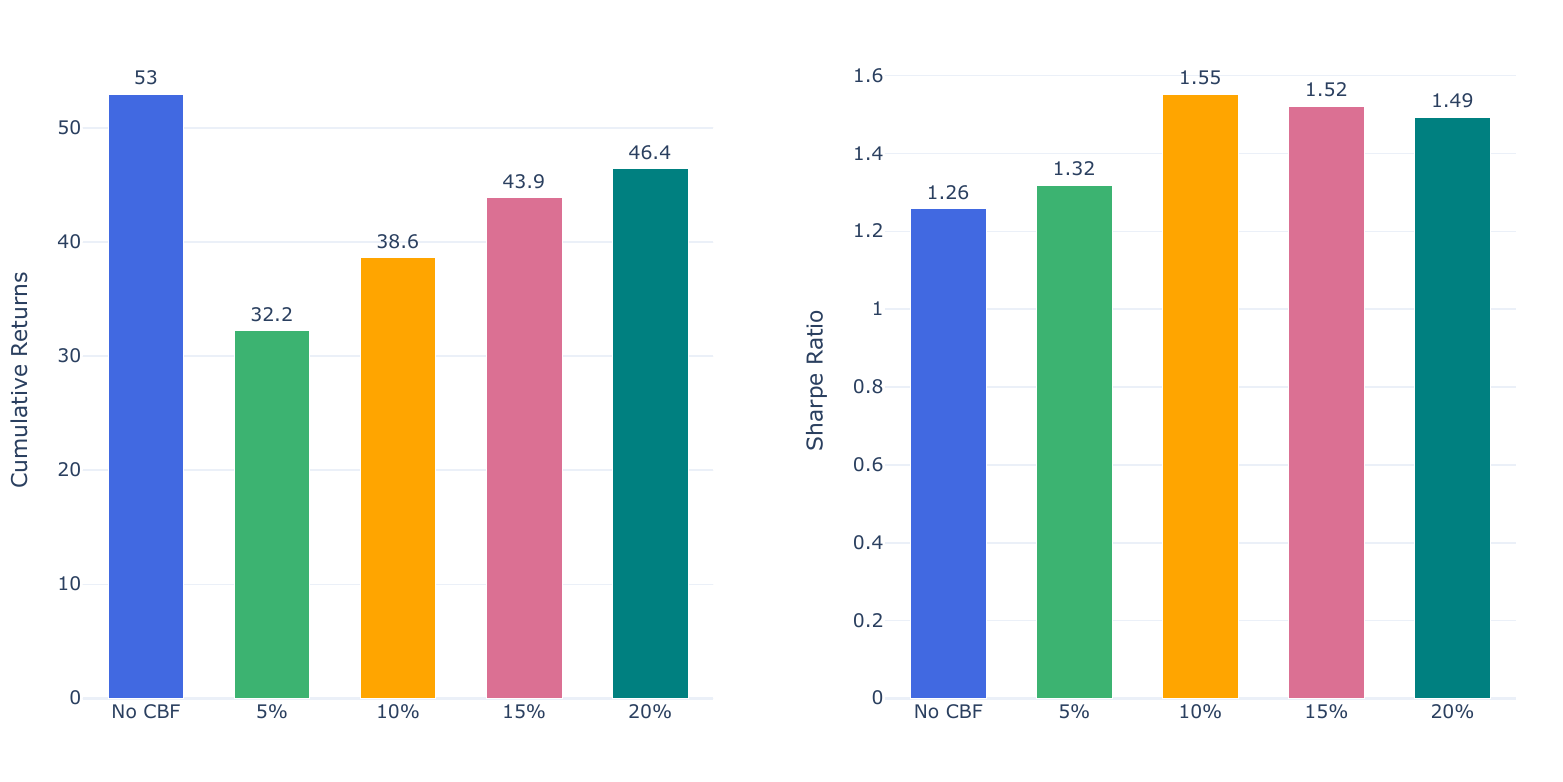}}
    \caption{Effect of varying the maximum drawdown limit in the control barrier function on cumulative return and Sharpe ratio.}
    \label{fig:drawdown_ablation}
  \end{center}
\end{figure}

To better understand the contribution of the main components of the proposed framework, we conduct a detailed ablation study using the DJI dataset. The first study examines the effect of varying the number of expert trajectories used to construct the barycenter representation. The second study analyzes the impact of the drawdown constraint imposed through the control barrier function. In both cases, we evaluate performance using cumulative return and the Sharpe ratio to capture profitability and risk-adjusted behavior. Figures~\ref{fig:expert_ablation} and \ref{fig:drawdown_ablation} present the ablation results, illustrating the effect of varying the number of expert trajectories used for barycenter construction and the drawdown limit in the control barrier function, respectively. In addition to these two studies, we carry out three further ablations to isolate the contribution of individual design choices in BRaG. Table~\ref{tab:barrycenter_ablation} compares the Sharpe-based expert-weighting scheme used in the barycenter against an equal-weighting alternative, to check whether the performance gains of BRaG stem from favoring stronger experts or merely from combining multiple experts. Table~\ref{tab:component_ablation} presents a component-wise ablation, allowing the contribution of every major module to be assessed individually rather than only in combination. Table~\ref{tab:expert_aggregation} compares different expert-aggregation strategies, in order to separate the benefit of using multiple experts at all from the benefit of the specific barycenter formulation adopted in this work. 

In the expert ablation study, we vary the number of expert trajectories used during training from one to five. When a single expert is used, it corresponds to the best-performing expert based on validation performance. For two, three, and four experts, the sets are formed by selecting the top-performing experts in descending order of performance. The final setting uses all five available experts. The results show a clear improvement in cumulative returns as more expert trajectories are incorporated. A similar trend is observed for the Sharpe ratio, which increases as the diversity of expert behavior grows. This suggests that aggregating multiple experts helps the model capture a richer set of trading patterns and reduces over-reliance on a single strategy. The marginal improvement from four to five experts is smaller, indicating that most gains are achieved once sufficient diversity is introduced.

The drawdown ablation study investigates the role of explicit risk control by varying the maximum allowable drawdown from $5\%$ to $20\%$ and by removing the control barrier function. Without any drawdown constraint, the model achieves high cumulative returns but exhibits weaker risk-adjusted performance. Introducing moderate drawdown limits results in a noticeable improvement in the Sharpe ratio, with the optimal balance observed at intermediate constraint levels. Very tight or very loose constraints result in suboptimal performance, either by limiting profitable opportunities or by allowing excessive risk. These results suggest that the control barrier function plays a crucial role in stabilizing learning and enhancing risk-adjusted returns when appropriately calibrated.

\begin{table}[!htp]\centering
\caption{Effect of expert-weighting schemes in the Wasserstein barycenter on the DJI dataset. Sharpe-based weighting assigns greater influence to experts with stronger risk-adjusted validation performance, whereas equal weighting assigns identical importance to all experts.} \label{tab:barrycenter_ablation}
\begin{tabular}{lcccc}\toprule
\textbf{Barycenter weighting scheme} &\textbf{Cumulative Return $\uparrow$} &\textbf{Sharpe Ratio $\uparrow$} &\textbf{Maximum Drawdown $\downarrow$} \\\midrule
\textbf{Sharpe-based weighting} &53.977 &1.472 &17.958 \\
\textbf{Equal weighting} &48.502 &1.281 &20.133 \\
\bottomrule
\end{tabular}
\end{table}

Table \ref{tab:barrycenter_ablation} presents the effect of the expert-weighting scheme used while constructing the Wasserstein barycenter. Two settings are compared here. The first is the Sharpe-based weighting used in the proposed BRaG framework, in which each expert is assigned a weight based on its risk-adjusted validation performance. The second is an equal-weighting scheme, in which every expert contributes equally to the barycenter regardless of how well it actually performed. This comparison is included to check whether the performance gains of BRaG come specifically from favoring better experts, or whether simply combining multiple experts is enough on its own. The results in Table 4 show that Sharpe-based weighting gives a clear advantage over equal weighting on all three metrics. Cumulative return drops from $53.977$ to $48.502$ when equal weighting is used, and the Sharpe ratio falls from $1.472$ to $1.281$. Maximum drawdown also worsens, rising from $17.958$ to $20.133$. This pattern suggests that when weaker experts are allowed to influence the pseudo-expert distribution as much as stronger ones, the resulting barycenter becomes a less reliable target for imitation learning. Greater emphasis on reliable experts produces a more useful pseudo-expert representation and improves both profitability and risk control.

\begin{table}[!htp]\centering
\caption{Component-wise ablation of BRaG on the DJI dataset. Each variant removes one major component from the full framework while retaining the remaining modules.} \label{tab:component_ablation}
\begin{tabular}{lcccc}\toprule
\textbf{Component removed} &\textbf{Cumulative Return $\uparrow$} &\textbf{Sharpe Ratio $\uparrow$} &\textbf{Maximum Drawdown $\downarrow$} \\\midrule
\textbf{GAIL Pretraining} &33.501 &1.245 &15.796 \\
\textbf{PPO Fine-tuning} &42.378 &1.335 &16.953 \\
\textbf{CBF} &52.951 &1.258 &25.874 \\
\textbf{None—Full BRaG} &53.977 &1.472 &17.958 \\
\bottomrule
\end{tabular}
\end{table}

Table \ref{tab:component_ablation} reports the component-wise ablation of BRaG on the DJI dataset, where each of the three main components, namely GAIL pretraining, PPO fine-tuning, and the control barrier function, is removed one at a time while keeping the remaining components intact. This experiment is carried out to understand how much each individual component contributes to the overall performance of the framework, since the full model combines several mechanisms together, and it is not obvious from the aggregate results alone which part is doing the heavy lifting. The results indicate that removing GAIL pretraining causes the sharpest fall in performance, with cumulative return dropping to $33.501$ and the Sharpe ratio to $1.245$. This shows that the barycenter-guided pretraining stage plays a central role in giving the policy a useful starting point before it is exposed to the true market reward. Removing PPO fine-tuning also hurts performance, though to a lesser extent, with a cumulative return of $42.378$ and a Sharpe ratio of $1.335$. This is expected because, without fine-tuning, the policy is limited to the behavior it can imitate from the pseudo-expert and never has the opportunity to improve further using actual portfolio returns. The CBF ablation tells a different kind of story. Removing the CBF actually results in a slightly higher cumulative return of $52.951$, close to the full model, but the Sharpe ratio drops to $1.258$, and the maximum drawdown increases sharply to $25.874$. This shows that the control barrier function does not contribute much to raw profitability, but it does most of the work in keeping the strategy's risk profile in check. Taken together, these three ablations show that the two learning-based components are primarily responsible for BRaG's return advantage, while the CBF is responsible for the risk-adjusted stability that distinguishes BRaG from other DRL baselines.

\begin{table}[!htp]\centering
\caption{Comparison of expert-demonstration aggregation strategies on the DJI dataset.} \label{tab:expert_aggregation}
\begin{tabular}{lcccc}\toprule
\textbf{Expert Aggregation Method} &\textbf{Cumulative Return $\uparrow$} &\textbf{Sharpe Ratio $\uparrow$} &\textbf{Maximum Drawdown $\downarrow$} \\\midrule
\textbf{Best Single Expert} &27.235 &1.205 &14.982 \\
\textbf{Uniform Expert Sampling} &46.445 &1.285 &20.181 \\
\textbf{Sharpe-weighted Sampling} &50.213 &1.388 &18.875 \\
\textbf{Full BRaG} &53.977 &1.472 &17.958 \\
\bottomrule
\end{tabular}
\end{table}

Table \ref{tab:expert_aggregation} compares different ways of aggregating expert demonstrations, ranging from using just a single best-performing expert to the full barycenter-based approach used in BRaG. Four settings are reported, namely the best single expert, uniform sampling across all experts, Sharpe-weighted sampling without any barycenter construction, and the full BRaG method. This comparison is meant to separate the value of using multiple experts at all from the value of the specific barycenter construction technique adopted in this work. The best single expert setting performs the weakest among the four, with a cumulative return of only $27.235$ and a Sharpe ratio of $1.205$, even though it has the lowest maximum drawdown of $14.982$. This is not surprising, since relying on a single expert restricts the policy to a single trading philosophy and prevents it from benefiting from complementary strategies from the other experts. Moving to uniform sampling improves the cumulative return considerably to $46.445$, confirming that combining multiple experts, even without any weighting, already yields a substantial benefit. Sharpe-weighted sampling further improves results, raising the cumulative return to $50.213$ and the Sharpe ratio to $1.388$, showing that giving greater weight to better-performing experts during sampling adds value on top of simple pooling. The full BRaG framework, which uses the barycenter construction along with Sharpe-based weighting, achieves the best results overall across all three metrics. This suggests that the barycenter is doing more than just weighting, since it also captures the underlying geometry of the expert distributions in a way that plain weighted sampling cannot fully replicate. Overall, the results in Table \ref{tab:expert_aggregation} indicate that both the multi-expert idea and the specific barycenter formulation contribute meaningfully, with the barycenter providing the final and most useful layer of improvement.

\end{document}